\documentclass[10pt, utf8, UTF8, twocolumn]{wlscirep}
\usepackage[fontsize=10pt]{fontsize}
\usepackage[colaction]{multicol}
\usepackage{amsmath}
\usepackage{mathrsfs}
\usepackage{float}
\usepackage{enumitem}
\usepackage{caption}
\usepackage{hyperref}
\usepackage{lineno}
\usepackage{wrapfig}
\usepackage{url}            
\usepackage{booktabs}       
\usepackage{amsfonts}       
\usepackage{nicefrac}       
\usepackage{microtype}      
\usepackage{graphicx}
\usepackage{color}
\usepackage{amssymb}
\usepackage{pifont}
\usepackage{subcaption}

 \newcommand{\blktitle}[1]{\vspace{2mm}\noindent\textbf{#1\quad}}

\newtheorem{definition}{Definition}
\newtheorem{theorem}{Theorem}

\DeclareMathAlphabet{\mathcal}{OMS}{cmsy}{m}{n}

\title{
\vspace{1.2cm}
\LARGE{Causally-informed Deep Learning towards Explainable and Generalizable Outcomes Prediction in Critical Care}
}

\author[1,$\dag$]{Yuxiao Cheng}
\author[1,$\dag$]{Xinxin Song}
\author[1]{Ziqian Wang}
\author[3]{Qin Zhong}
\author[1,2,*]{Qionghai Dai}
\author[3,*]{Kunlun He}
\author[1,2,4,*]{Jinli Suo}
\affil[1]{Department of Automation, Tsinghua University, Beijing, China}
\affil[2]{Institute of Brain and Cognitive Sciences, Tsinghua University, Beijing, China}
\affil[3]{Chinese PLA General Hospital, Beijing, China}
\affil[4]{Shanghai Artificial Intelligence Laboratory, Shanghai, China}
\affil[*]{qhdai@tsinghua.edu.cn; kunlunhe@plagh.org; jlsuo@tsinghua.edu.cn}
\affil[$\dag$]{these authors contributed equally to this work}

\begin{abstract}
\textbf{
Recent advances in deep learning (DL) have prompted the development of high-performing early warning score (EWS) systems, predicting clinical deteriorations such as acute kidney injury, acute myocardial infarction, or circulatory failure. DL models have proven to be powerful tools for various tasks but come with the cost of lacking interpretability and limited generalizability, hindering their clinical applications. To develop a practical EWS system applicable to various outcomes, we propose causally-informed explainable early prediction model, which leverages causal discovery to identify the underlying causal relationships of prediction and thus owns two unique advantages: demonstrating the explicit interpretation of the prediction while exhibiting decent performance when applied to unfamiliar environments. Benefiting from these features, our approach achieves superior accuracy for 6 different critical deteriorations and achieves better generalizability across different patient groups, compared to various baseline algorithms. Besides, we provide explicit causal pathways to serve as references for assistant clinical diagnosis and potential interventions. The proposed approach enhances the practical application of deep learning in various medical scenarios.
}

\end{abstract}

\begin{document}

\captionsetup[figure]{labelfont={bf},name={Fig.},labelsep=period}
\twocolumn[
\begin{@twocolumnfalse}
\maketitle
\end{@twocolumnfalse}
]

\begin{figure*}[t]
\centering
\includegraphics[width=\linewidth, trim={{5.75in 12.6in 5.75in 2in}}, clip]{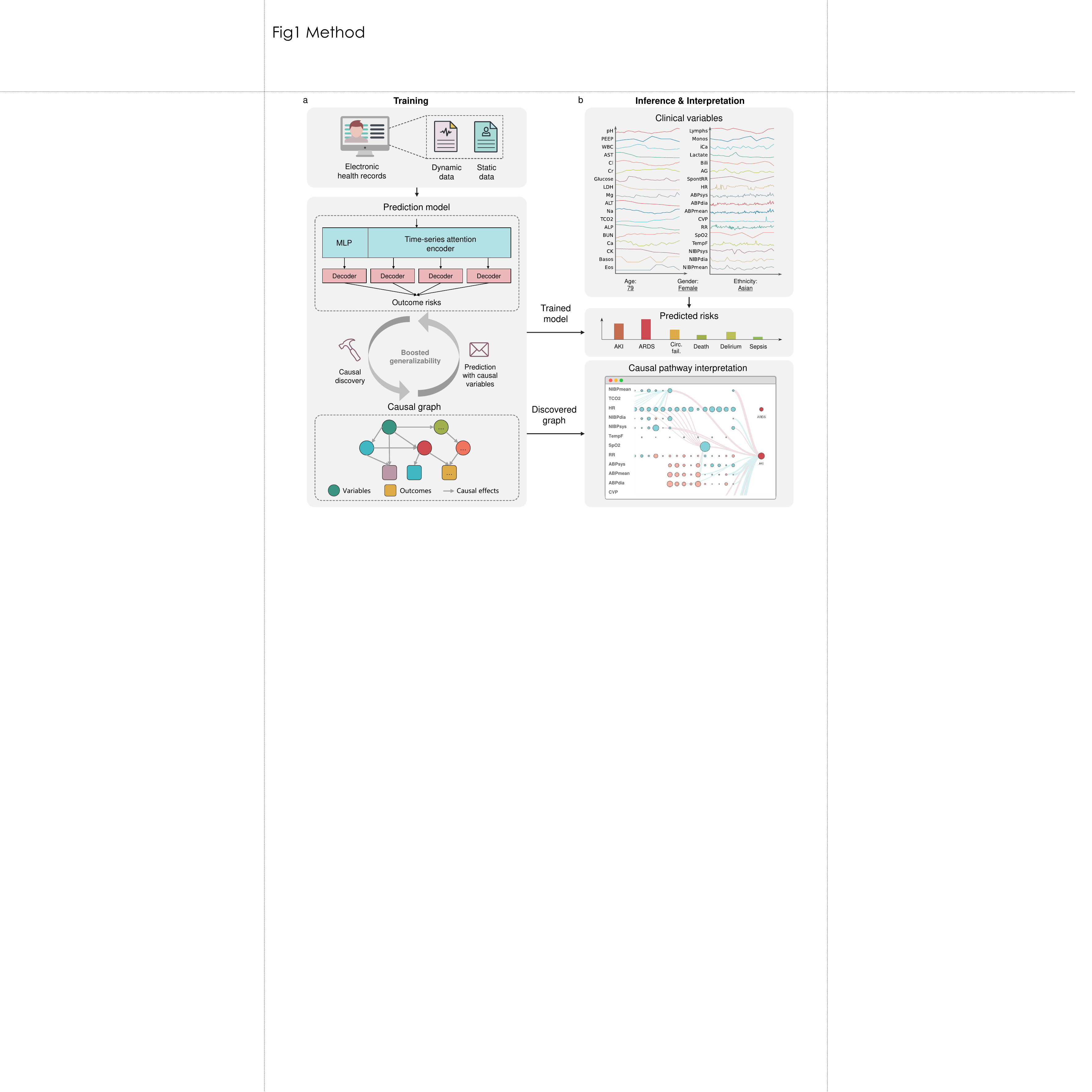}
\caption{
{\bf The overall architecture of the proposed approach. }
\textbf{a,} The cDEEP model is learned from a large collection of electronic health records, including dynamic clinical monitoring data and the patients' static demographic variables.
During training, cDEEP optimizes the causal graphs and neural networks iteratively to unveil direct causal effects, enhancing the model's generalizability. 
\textbf{b,} In the inference process,
cDEEP enhances the interpretability of deep learning models by providing explicit causal pathways for predictions, enabling clinicians to identify actionable variables throughout the chain. 
The detailed architecture and network design are provided in Extended Data Fig. \ref{fig:train}.}
\label{fig:method}
\end{figure*}

\begin{figure*}[p]
\centering
\includegraphics[width=\linewidth, trim={{5.75in 6.2in 5.75in 2.0in}}, clip]{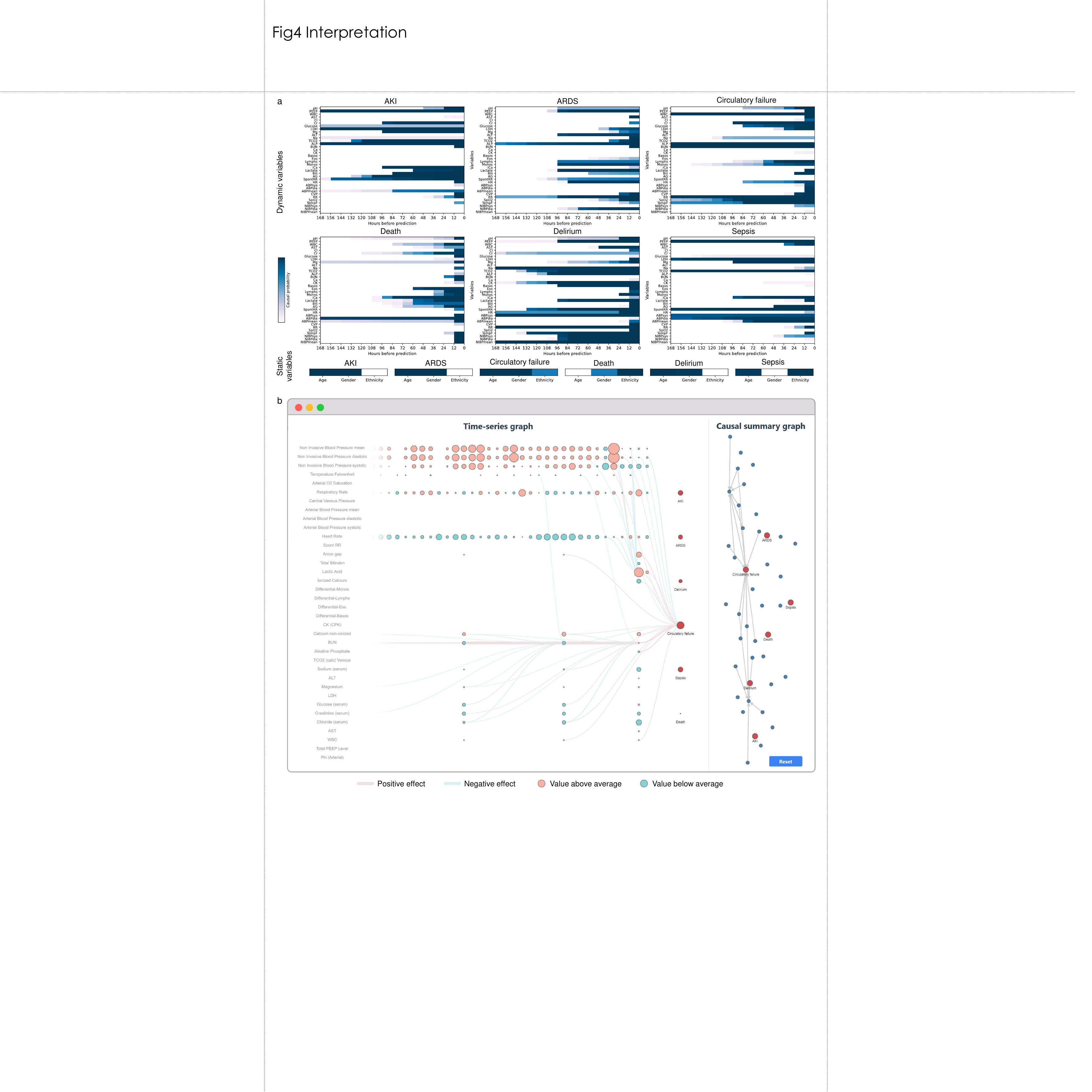}
\caption{ {\bf Demonstration of cDEEP's interpretable predictions.} Caption continues on the next page.
}
\label{fig:intp}
\end{figure*}

\begin{figure*}[t]
\centering
\addtocounter{figure}{-1}
\caption{ {\bf Demonstration of cDEEP's interpretable predictions.} 
\textbf{a,} Illustration of the causal probability matrix that shows the probability of each variable at time point $t$ being a direct cause of the outcome. Details of the causal probability matrix are described in \textit{Methods}. 
\textbf{b,} Visualization of the causal pathways and controlled direct effect (CDE) values provided by our web-based visualization tool. This tool enables clinicians to interact with the nodes of the causal graph, allowing them to examine the CDE values and causal pathways, thereby clarifying the underlying relationships and the overall impact of each variable on the outcomes. 
Each circle in the graph signifies a variable, with color coding indicating the polarity of deviation from the average (red for above average and blue for below average), while the size of the circle reflects the magnitude of the deviation. Causal pathways are represented by arrows pointing from the cause to the effect, with the thickness of the arrows corresponding to the CDE values, and the color of the arrows indicating the polarity of the causal effect (red for positive and blue for negative). 
We deployed this tool on \url{https://cdeep.icu/} and
see \textit{Supplements C.2} for user guidance.
}
\end{figure*}

\begin{table*}[t]
    \centering
    \small
    \tabcolsep=0.1cm
    \caption{Description of the patient population.}
    \label{tab:desc}
    \begin{tabular}{p{7.5cm}p{3cm}p{3cm}p{3cm}}
    \toprule
    \textbf{Description} & \textbf{In-dist.} & \textbf{Out-of-dist.} &  \textbf{Overall} \\ \midrule
    Unique patients, no. & 135810 & 51617 & 187427 \\ 
    Unique admissions, no. & 192013 & 71012 & 263025 \\ 
    Age, median years & 59 & 83 & 66 \\ 
    Gender, male, \% of total admissions & 56.974  & 48.042  & 54.515  \\ 
    AKI present, \% of total admissions & 34.67\% (66567) & 42.84\% (30422) & 36.87\% (96989) \\ 
    ARDS present, \% of total admissions & 8.05\% (15454) & 7.15\% (5076) & 7.81\% (20530) \\ 
    Circulatory failure present, \% of total admissions & 25.92\% (49779) & 26.96\% (19146)	& 26.20\% (68925) \\ 
    Death present, \% of total admissions & 4.77\% (9154) & 8.17\% (5803) & 5.69\% (14957) \\ 
    Delirium present, \% of total admissions & 6.80\% (13063) & 9.36\% (6648) & 7.49\% (19711) \\ 
    Sepsis present, \% of total admissions & 7.62\% (14622) & 8.30\% (5895) & 7.80\% (20517) \\ 
    
    \bottomrule
    \end{tabular}
\end{table*}

\begin{table*}[t]
    \centering
    \small
    \tabcolsep=0.05cm
    \footnotesize
    \caption{List of all input variables for outcomes prediction, with corresponding abbreviations shown in bold.}
    \label{tab:var}
    \begin{tabular}{llllll}
    \toprule
    \multicolumn{6}{c}{\textit{\textbf{Dynamic Variables}}}                                                                                                                      \\ \midrule
    \textbf{pH}      & PH (Arterial)                        & \textbf{PEEP}    & Positive End-Expiratory Pressure      & \textbf{WBC}       & White Blood Cell                   \\
    \textbf{AST}     & Aspartate Aminotransferase           & \textbf{Cl}      & Chloride (serum)                      & \textbf{Cr}        & Creatinine (serum)                 \\
    \textbf{Glucose} & Glucose (serum)                      & \textbf{LDH}     & Lactate Dehydrogenase                 & \textbf{Mg}        & Magnesium                          \\
    \textbf{ALT}     & Alanine Transaminase                 & \textbf{Na}      & Sodium (serum)                        & \textbf{TCO2}      & Total Carbon Dioxide Venous \\
    \textbf{ALP}     & Alkaline Phosphate                   & \textbf{BUN}     & Blood Urea Nitrogen                   & \textbf{Ca}        & Calcium non-ionized                \\
    \textbf{CK}      & Creatine Kinase                      & \textbf{Basos}   & Differential-Basos                    & \textbf{Eos}       & Differential-Eos                   \\
    \textbf{Lymphs}  & Differential-Lymphs                  & \textbf{Monos}   & Differential-Monos                    & \textbf{iCa}       & Ionized Calcium                    \\
    \textbf{Lactate} & Lactic Acid                          & \textbf{Bili}    & Total Bilirubin                       & \textbf{AG}        & Anion gap                          \\
    \textbf{SpontRR} & Spontaneous Respiratory Rate         & \textbf{HR}      & Heart Rate                            & \textbf{ABPsys}    & Ambulatory Blood Pressure systolic \\
    \textbf{ABPdia}  & Ambulatory Blood Pressure diastolic  & \textbf{ABPmean} & Ambulatory Blood Pressure mean        & \textbf{CVP}       & Central Venous Pressure            \\
    \textbf{RR}      & Respiratory Rate                     & \textbf{SpO2}    & Arterial O2 Saturation                & \textbf{TempF}     & Temperature Fahrenheit             \\
    \textbf{NIBPsys} & Non Invasive Blood Pressure systolic & \textbf{NIBPdia} & Non Invasive Blood Pressure diastolic & \textbf{NIBPmean}  & Non Invasive Blood Pressure mean   \\ \bottomrule
    \multicolumn{6}{c}{\textit{\textbf{Static Variables}}}                                                                                                                       \\ \midrule
    \textbf{Age}     & Age                                  & \textbf{Gender}  & Gender                                & \textbf{Ethnicity} & Ethnicity                          \\ \bottomrule
    \end{tabular}
\end{table*}

\noindent The rapid development of deep learning technologies has yielded some powerful early warning score (EWS) tools for predicting critical clinical deterioration events, e.g., acute kidney injury, acute myocardial infarction, and circulatory failure. 
Although achieving promising performance in various tasks, deep learning models are faced with multiple limitations that significantly hinder their practical applications, including low interpretability and limited generalizability\cite{prosperiCausalInferenceCounterfactual2020, vollmerMachineLearningArtificial2020, shamoutMachineLearningClinical2021, yangUnboxBlackboxMedical2022}.

Interpretability and generalizability are especially important for a clinically practical EWS system because physicians intend to trust a prediction with explicit inference mechanisms as well as consistent accuracy in different environments. 
Researchers have made a lot of attempts to build high-performance deep neural networks with high interpretability and generalizability. The techniques enhancing interpretability involve post hoc methods such as GradCAM, LIME, or SHAP\cite{lauritsenExplainableArtificialIntelligence2020, caicedo-torresISeeUVisuallyInterpretable2019a, steyaertMultimodalDataFusion2023}, attention mechanism\cite{ethayarajhAttentionFlowsAre2021}, decision trees\cite{yanInterpretableMortalityPrediction2020} and knowledge distillation\cite{yangUnboxBlackboxMedical2022}. 
For better generalizability, various strategies have been proposed,  including dropout\cite{srivastavaDropoutSimpleWay2014}, weight decay\cite{kroghSimpleWeightDecay1991}, or more complex approaches such as invariant risk minimization (IRM), GroupDRO, or VREx\cite{arjovskyInvariantRiskMinimization2020a, sagawa*DistributionallyRobustNeural2019, kruegerOutDistributionGeneralizationRisk2021}, etc. Despite the efforts and progress in the above two respective directions, 
in the field of medical AI, the interpretability is often constrained to variable-to-outcome explanations through post-hoc methods like SHAP \cite{lauritsenExplainableArtificialIntelligence2020,thorsen-meyerDynamicExplainableMachine2020}. These methods fail to show underlying causal structures, and thus only provide limited information for potential clinical intervention. Additionally, most algorithms are not specially designed to boost generalizability by considering the intrinsic relationships among the clinical variables and outcomes\cite{hylandEarlyPredictionCirculatory2020, batesHarnessingAISepsis2022, demirjianPredictiveAccuracyPerioperative2022, thanMachineLearningPredict2019}.

Causal models are exactly what can address the above issues since causality is stable and interpretable\cite{castroCausalityMattersMedical2020, rungeCausalInferenceTime2023, pearlCausalityModelsReasoning2009, vowelsDyaDagsSurvey2021, shenStableLearningSample2020}. 
For example, it is well-established among clinicians that elevated insulin levels lead to a reduction in blood glucose levels, i.e., insulin causes decreased glucose. Consequently, predicting a decrease in blood glucose based on an observation of high insulin levels is both interpretable and consistent across nearly all individuals and clinical settings.
Manually building the complex and entangled causal relationship behind numerous diseases is extremely difficult and sometimes even impossible. However, recent advancements in DL-based causal discovery algorithms have made it possible to construct intricate causal models by analyzing vast datasets\cite{moraffahCausalInferenceTime2021, chengCUTSNeuralCausal2023, chengCUTSHighDimensionalCausal2024, lagemannDeepLearningCausal2023}, providing a promising approach to developing reliable and practical EWS.

This study presents the causally-informed Dynamic Explainable Early Prediction (cDEEP), a novel method that employs causal discovery to identify stable associations between the clinical variables of ICU patients and their outcomes. 
Specifically, the causal graph captures causal variables that are important to each outcome, boosting the generalizability of the model by only including causal variables. Moreover, it explores causal associations among dynamic variables, which facilitates identifying the inference path of the prediction explicitly and enhances the interpretability further. The graph would empower clinicians to identify actionable variables throughout the chain, thereby facilitating more effective interventions that may reduce outcome risks. 

The findings demonstrate that cDEEP effectively reveals the intricate relationships among numerous variables, ruling out spurious variables and constructing stable causal graphs. Utilizing the identified causal graph, cDEEP is capable of generating interpretable and generalizable predictions for all targeted outcomes with a single model (including AKI, ARDS, circulatory failure, death, delirium, and sepsis and can be extended further if provided with proper training data), thus improving the applicability of deep learning in various out-of-distribution testing data.


\section*{Results}

\blktitle{Data.} We used two large electronic health record (EHR) databases---the Medical Information Mart for Intensive Care (MIMIC) IV \cite{johnsonMIMICIVFreelyAccessible2023} and the eICU Collaborative Research Database \cite{pollardEICUCollaborativeResearch2018}. Six medical outcomes were identified: acute kidney injury (AKI), acute respiratory distress syndrome (ARDS), circulatory failure, death, delirium, and sepsis. As shown in Extended Fig. \ref{fig:dataalloc}a,
all patients under or at the age of $75$ were randomly split into training, validation, and in-distribution testing sets (80\%, 10\%, and 10\%), while the out-of-distribution testing sets comprised patients aged 76 and above. 
Additionally, we partitioned the out-of-distribution testing sets according to admission time for additional experiments aimed at assessing generalizability.
Shown in Extended Data Fig. \ref{fig:dataalloc}b, dynamic data from the databases are converted to temporally structured sequences, and the predictions are made dynamically at each time point. 
Patient population statistics are shown in Tab. \ref{tab:desc}, and all included variables are listed in Tab. \ref{tab:var}.
For further information regarding data preprocessing, please consult the \textit{Methods} section.

\blktitle{Overview of cDEEP.} Current deep-learning-based approaches, though excel at making accurate decisions, primarily focus on establishing correlations rather than causations, leading to unreliable decisions due to a lack of fundamental interpretability and generalizability (refer to \textit{Supplements A} for more detailed literature reviews). As a result, some recent literature delves into causal models, among which the most examined framework is Pearl's structural causal model (SCM), with variables depicted as nodes and causal influences indicated by directed edges. Recently, various causal discovery approaches have been proposed to build SCMs with observational data\cite{moraffahCausalInferenceTime2021, chengCUTSNeuralCausal2023, chengCUTSHighDimensionalCausal2024, lagemannDeepLearningCausal2023}, helping to develop reliable and practical deep learning models.

The proposed cDEEP, demonstrates a novel early warning system scheme that integrates causal discovery with deep learning to predict six critical care outcomes. As illustrated in Fig. \ref{fig:method}a, after collecting dynamic and static data from the EHR databases, we trained the deep learning model to assess the risk of a patient experiencing a specific outcome within the next 24 hours.
The model combines an encoder-decoder neural network architecture with a causal discovery process to identify and utilize the causal relationships among the input variables and outcomes.  
During clinical applications illustrated in Fig. \ref{fig:method}b, cDEEP takes advantage of the learned causal graph and the prediction model to estimate the risks of six outcomes and offers explicit causal pathways for high interpretability.

We train the model by iterating two key steps: i) optimizing the prediction neural network given a fixed causal graph and ii) refining the causal graph based on the current neural network. 
The prediction model is implemented with an encoder-decoder neural network: a multi-layer perceptron (MLP) and a time-series attention network encode static and dynamic data respectively; a series of MLPs to decode the future values of variables or risks for outcomes. 
The causal graph is implemented with a probability matrix and is relaxed as a continuous form to be optimized in the loss function.
As for the causal graph, we split the causal graph to be discovered into variable-to-outcome (V2O) and variable-to-variable (V2V) graphs to explore the causal pathways behind the neural network's prediction. Besides, the graph identifies direct causal variable contributing to the clinical outcome, which enhances the generalizability of the prediction model. 
The formulation and steps of the training process are comprehensively described in the \textit{Methods} section and illustrated in Extended Data Fig. \ref{fig:train}.

\begin{figure*}[t!]
\centering
\includegraphics[width=\linewidth, trim={{5.75in 11.1in 5.75in 2in}}, clip]{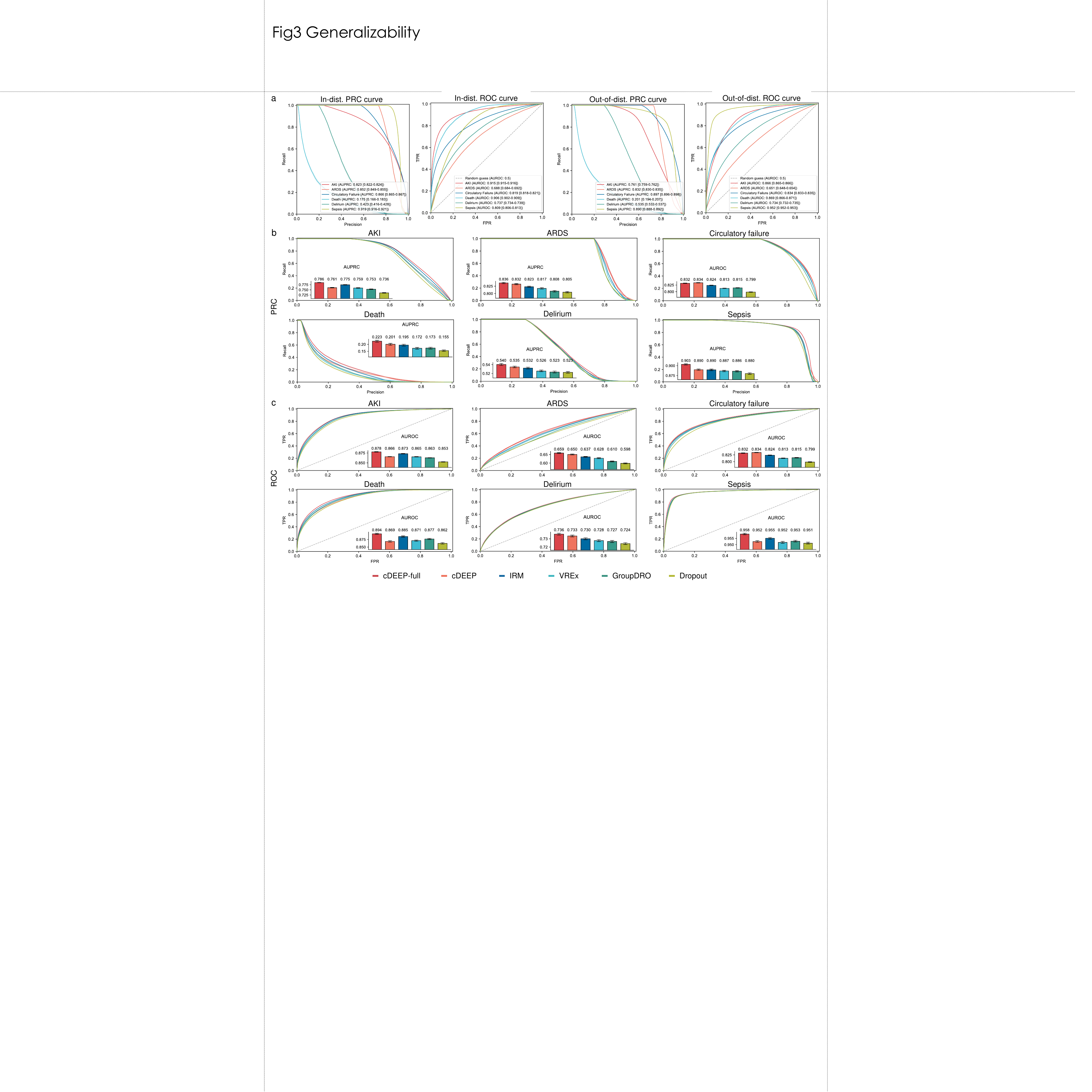}
\caption{
{\bf cDEEP's advantageous performance on out-of-distribution testing datasets.}
\textbf{a,} Illustration of receiver operating characteristic curve (ROC) and precision-recall curve (PRC) on in-distribution and out-of-distribution testing data.
\textbf{b-c,} A comparative analysis of the performance on out-of-distribution data against existing generalizable AI methodologies in terms of ROC and PRC, respectively. cDEEP-full refers to our model that inputs all variables, while cDEEP inputs only causal variables. 
}
\label{fig:gen}
\end{figure*}

\blktitle{High prediction accuracy.} cDEEP can predict multiple outcomes with a single neural network and achieves high predictive accuracy in the testing set, with the results illustrated in Fig. \ref{fig:gen}a. 

cDEEP performs quite well in the prediction of AKI, Circulation failure, Delirium, and Sepsis. Taking AKI as an example, we achieve an area under the receiver operating characteristic curve (AUROC) of 0.915, with 95\% confidence interval (CI) over 100 cross-validation folds being [0.915, 0.916]; area under the precision-recall curve (AUPRC) is 0.823 (CI [0.822, 0.824]).

The prediction accuracies for some outcomes are affected by lacking or imbalanced labeling (see Supplements Tab. S1). 
For example, based on the Berlin definition, ARDS is diagnosed when the PaO2/FiO2 ratio is smaller than 300, so the prediction windows with no PaO2/FiO2 ratio were excluded from the training set, leading to 94.4\% prediction points being disregarded,   
resulting in a relatively low AUROC at 0.688 (CI [0.684-0.692]). Despite the insufficient training data, the percentage of ARDS positive prediction points among the left 5.6\% training data is high, we achieve a decent AUPRC 0.852 (CI [0.849-0.855]), validating cDEEP's prediction capability. 
The problem of data imbalance is severe for death---only 0.9\% death-positive prediction points in the database, which hampers the prediction accuracy. Although high AUROC is achieved (0.906, CI [0.902-0.909]), in the scenarios with imbalance data, AUPRC is a better metric for model evaluation and the AUPRC for death prediction is only 0.194 (CI [0.185-0.204]).

Calibration curves for all six outcomes are shown in \textit{Supplements Fig. S2}. By performing isotonic regression, the calibration performance is only slightly improved (here the Brier score decreases by 0.0004). 
These results show that cDEEP is already well calibrated, indicating that the predicted probabilities are consistent with the actual counterpart.

\begin{figure}[t!]
\centering
\includegraphics[width=\linewidth, trim={{5.75in 15in 11.8in 2in}}, clip]{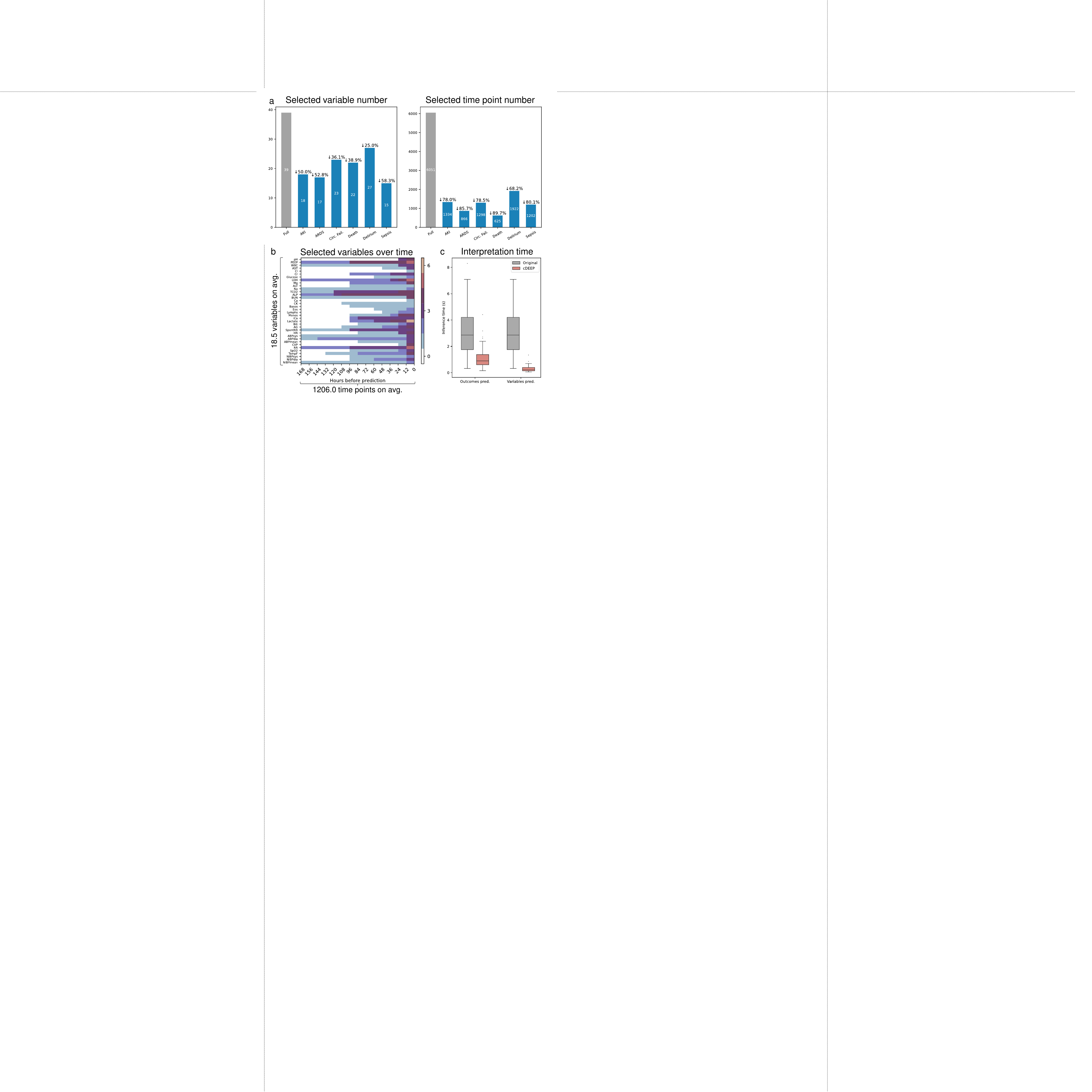}
\caption{
{\bf Acceleration for quantitative interpretation.}
\textbf{a,} The decreased number of required input variables (left) and time points (right) for cDEEP to predict each clinical outcome for the acceleration of quantitative interpretation. 
\textbf{b, } Visualization of the selected variables over time.
\textbf{c,} Average computation time for a full quantitative interpretation before and after acceleration.
}
\label{fig:dataint}
\end{figure}

\blktitle{Qualitative interpretation.} Causal models possess inherent interpretability that allows for the elucidation of neural network decisions by analyzing the parent nodes within a causal graph. It follows logically that the parents in the causal graph are responsible for the outcomes of interest, as illustrated in Fig. \ref{fig:intp}a. 
By integrating V2V and V2O graphs, cDEEP obtains a comprehensive causal graph encompassing both variables and clinical outcomes. For a visual representation of the constructed causal graph, please refer to Extended Data Fig. \ref{fig:graph}.
This combined causal graph enables the identification of the causal pathways that illustrate the influence of each variable. Taking the path chloride $\rightarrow$ BUN $\rightarrow$ circulatory failure as an example, we explicitly show the inference philosophy behind the cDEEP's prediction of circulatory failure's occurrence based on observation of chloride, through the intermediary variables BUN.

Providing causal pathways allows clinicians to pinpoint key intervention opportunities following the path, enabling targeted interventions. For instance, when a critical vital sign is identified as the direct cause of an outcome and is challenging to address directly, we can trace back through the causal pathway to discover actionable causes related to that vital sign. This kind of insight is beneficial in clinical practice. 

The selected causal variables and the constructed causal pathways by our algorithm align with the known medical mechanisms and causes of critical conditions.
i) 
For AKI, creatinine, a diagnostic marker per KDIGO guidelines\cite{KDIGO2012}, reflects renal dysfunction; elevated PEEP increases intrathoracic pressure, reducing renal perfusion\cite{santacruzHighLowPositive2021}; rising lactate signals tissue hypoxia, while lactate dehydrogenase indicates renal and systemic injury\cite{basilePathophysiologyAcuteKidney2012}. 
ii) 
For ARDS, ALT and ALP suggest liver involvement in the inflammatory process, which affects ARDS severity\cite{matuschakOrganInteractionsAdult1988, herreroLiverLungInteractions2020}. Heart Rate (HR) and Respiratory Rate (RR) increase as compensatory responses to hypoxemia\cite{trillo-alvarezAcuteLungInjury2011}. PEEP aids in alveolar recruitment, improving oxygenation\cite{writinggroupforthealveolarrecruitmentforacuterespiratorydistresssyndrometrialartinvestigatorsEffectLungRecruitment2017}. Peripheral Oxygen Saturation (SpO2) reflects tissue oxygenation and decreases in ARDS\cite{rivielloHospitalIncidenceOutcomes2016, wickPulseOximetryDiagnosis2022}.
iii) 
In circulatory failure, low pH indicates metabolic acidosis\cite{morrisMetabolicAcidosisCritically2008}; reduced respiratory rate reflects late-stage respiratory depression\cite{marikHemodynamicParametersGuide2011}. 
iv) 
Mortality links to immune dysfunction and organ damage; abnormal lymphocytes signal adaptive immunity loss\cite{liImmuneDysfunctionLeads2020, adriePersistentLymphopeniaRisk2017}. 
v)
For delirium, an increased anion gap and heightened respiratory rate indicate metabolic acidosis and systemic inflammation, impairing cerebral metabolism; elevated PEEP exacerbates hypoxia by reducing cerebral venous return\cite{gouveabogossianEffectIncreasedPositive2023}. 
vi)
For sepsis, altered magnesium and anion gap levels indicate metabolic dysfunction\cite{velissarisHypomagnesemiaCriticallyIll2015}; 
cytokines like IL-6 mediate vasodilation, impairing perfusion to vital organs\cite{aliyuInterleukin6CytokineOverview2022}. 
These selections are reasonable and provide insights into the possible mechanisms driving these critical conditions. For more detailed interpretations, please refer to the \textit{Supplements C.1}.

\blktitle{Quantitative interpretation and its acceleration.} Nevertheless, while the analysis mentioned above qualitatively identifies the sources of the outcomes of interest, it does not quantitatively show the strength of the effects.
Additionally, the causal graphs offer a global interpretation of predictions across its entire dataset, leaving the individual interpretation (i.e., specifically how the test results of a particular patient affect the prediction) unexplored. To answer these questions, we further calculate the controlled direct effect (CDE) for each variable concerning the outcomes.  CDE serves as a quantitative assessment of the causal influence of a variable on the outcomes, determined by altering the variable and observing the resultant changes in predictions. 

Calculating the CDE values for all variables across all outcomes is necessary to effectively illustrate the causal pathways, a notably time-intensive process. 
Our proposed method, cDEEP, mitigates this challenge by limiting the input variables to only the causal variables, thereby reducing the computation time.
In this manner, 39 input variables (detailed in Tab. \ref{tab:var}) are decreased to an average of 18.5, significantly lowering data intensity for practical use. In application, cDEEP takes only the direct causes of the outcomes as input, which can be easily gathered using a simple applet or website, potentially reducing the need for frequent physiological tests or examinations.
Specifically, as shown in Fig. \ref{fig:dataint}c, after building a variable-to-outcome (V2O) graph, 18, 17, 23, 22, 27, and 25 direct causes are identified as critical inputs for AKI, ARDS, circulatory failure, death, delirium, and sepsis, respectively. 
Moreover, the distillation of input variables occurs not only across the variable axis but also along the time axis. 
We incorporate a cumulative window graph (see \textit{Methods}) to minimize the time lags of causal dependencies, thereby further reducing data intensity.
As a result, while maintaining a high level of prediction accuracy, the total number of input features to the neural network has been decreased by 80.1\%, as shown in Fig. \ref{fig:dataint}a-b. 

Additionally, we propose an accelerated inference technique, which allows for updating only a subset of the neural network's hidden layers following the perturbation of a variable, rather than conducting a complete inference. This methodology is elaborated in the \textit{Methods} section. Experiments in Fig. \ref{fig:dataint}c show that this acceleration technique decreases the computation time for constructing the V2O and V2V graphs by 63.1\% and 90.5\%, respectively.  

To better demonstrate the decision rationale behind cDEEP's outcome prediction, we designed a web-based tool to visualize the causal pathways alongside the corresponding CDE values. The tool features a user-friendly interface, allowing clinicians to interact with the causal graph to view the CDE values and the causal pathways. 
The user guide of this tool is provided in \textit{Supplements C.2}.
An example demonstrating the prediction of a patient undertaking circulatory failure in the following 24 hours is shown in Fig. \ref{fig:intp}b, with more examples provided in the \textit{Supplements C.3}.

\blktitle{Generalizability.} The theoretical foundation for the generalizability of cDEEP is established through its causal discovery process. 
As shown in Theorem \ref{theorem1}, by assuming that distribution shifts arise from interventions on various variables, cDEEP exclusively utilizes causal variables as input and thus can sustain consistent predictions despite such interventions, since the direct causal influences on the outcomes remain unaffected.

To assess the generalizability of our approach, we created out-of-distribution testing data by splitting the datasets by patients' ages, i.e., training on patients with age $\leq 75$ and testing on patients with age $\geq 76$.
We show that cDEEP achieves high prediction accuracy in out-of-distribution testing data, shown in Fig. \ref{fig:gen}a. 

Existing generalizable AI approaches, e.g., IRM\cite{arjovskyInvariantRiskMinimization2020a}, GroupDRO\cite{sagawa*DistributionallyRobustNeural2019}, and VREx\cite{kruegerOutDistributionGeneralizationRisk2021}, base the predictions on all available input variables while cDEEP only uses the direct causal variables.
Theoretically, Theorem \ref{theorem1} shows that by inputting only causal variables, cDEEP is the optimal worst-case generalizable model. Yet this is not to say that the non-causal variables serve no purpose in improving the prediction accuracy. Actually, the non-causal variables may contain some information that is not fully captured by the causal variables. As a result, direct comparisons between cDEEP and these methods may not be entirely fair. 
To further experimentally substantiate the generalizability of cDEEP, we finetuned cDEEP on all available input variables, which we refer to as ``cDEEP-full''. The comparisons of ``cDEEP-full'' (which inputs all variables) as well as ``cDEEP''  (which inputs causal variables) with the existing generalizable AI methodologies---IRM, VREx, GroupDRO, and Dropout---are shown in Fig. \ref{fig:gen}b-c and \textit{Supplements Tab. S3-4}. It is observed that cDEEP achieved competing or superior performance in most cases in terms of AUROC and AUPRC already, even with significantly fewer input variables.
On the other hand, the performance of cDEEP-full is even better than cDEEP, beating baseline methods in most cases, which further validates the superiority of cDEEP in terms of generalizability. Furthermore, 
experiments on robustness to noise
also indicate that cDEEP and cDEEP-full outperform baseline methods in most cases across various noise levels, as presented in \textit{Supplements Fig. S5}.

We also created out-of-distribution testing data by splitting the datasets by admitting time, i.e., training on patients admitted before 2014 and testing on patients admitted after. The results show similar conclusions as the age-based split, which is shown in \textit{Supplements Tab. S5-6}.

\section*{Discussions}


\noindent This study underscores the integration of deep learning with causal discovery for predicting clinical outcomes. Our methodology, termed cDEEP, focuses on causal relationships rather than mere correlations to uncover the small number of direct causal variables for outcomes and the inferring path for outcome prediction. Such a shift from association-based to causality-based models fills a significant void in existing deep learning-based EWS systems, which often struggle with low generalizability and limited interpretability, and lay a foundation for practical AI-based EWS tools.

The generalizability of cDEEP represents a significant advancement in the realm of medical artificial intelligence \cite{hylandEarlyPredictionCirculatory2020, batesHarnessingAISepsis2022, demirjianPredictiveAccuracyPerioperative2022, thanMachineLearningPredict2019}.
By constructing stable causal graphs across diverse environments and populations, cDEEP showcases robust performance even in out-of-distribution testing scenarios. This feature is essential for deploying Early Warning Systems (EWS) in various clinical contexts, where patient demographics and conditions may vary considerably from the training data. 
cDEEP's ability to sustain high accuracy under such circumstances highlights its potential as a dependable tool for early prediction in critical care settings. 
Our advantageous generalizability is fundamentally different from the existing AI-driven medical tools, which adjust the model structure or parameters to balance the capacity and generalizability and conduct external experiments to validate their design, instead of offering a more thorough methodology to enhance generalizability.

The second distinguishing feature of cDEEP is its interpretability\cite{thorsen-meyerDynamicExplainableMachine2020,cohenLegalEthicalConcerns2017}. By elucidating explicit causal relationships between variables and outcomes, cDEEP enables clinicians to gain a clear understanding of how each variable impacts the prediction. The visualization tool developed in this study further improves the interpretability of cDEEP by allowing clinicians to investigate the causal pathways and the controlled direct effects of each variable on the outcomes. 
Moreover, cDEEP accelerates the calculation of causal pathways by reducing input variables to essential causal variables, significantly decreasing the computation time required for interpretation.
This level of transparency is vital for fostering trust in AI-based clinical decision support systems and facilitating their practical use. 

Our methodology of achieving high interpretability via causal discovery substantially advances the previous studies in explainable early warning systems in critical care settings \cite{lauritsenExplainableArtificialIntelligence2020}. For example, Meyer et al. \cite{thorsen-meyerDynamicExplainableMachine2020} have developed a dynamic explainable AI model for predicting 90-day mortality rates among ICU patients and conducted post-hoc interpretability analysis by calculating the SHAP values. However, they only offer a localized attribution between variables and outcomes, and the SHAP-based interpretation method only focuses solely on the impact of input variables on the outcomes and leaves the variable-to-variable impact unexplored. 
As discussed in their paper, clinicians may struggle to pinpoint effective risk-reduction strategies without a comprehensive view of how variables interact and affect one another. 
Conversely, the proposed model encompasses the entire causal pathway (as illustrated in Fig. \ref{fig:intp}b) facilitating a more extensive analysis,  revealing multiple potential intervention points. 
Addressing an arbitrary variable within this chain could enhance clinicians' decision-making processes and improve patient outcomes.

Currently, cDEEP focuses primarily on structured clinical data. Expanding to incorporate other data modalities, such as medical imaging, could further enhance the model's diagnostic capability and generalizability across diverse clinical scenarios. Additionally, we aim to develop an extended version of cDEEP that factors in medical interventions, enabling the system to suggest potentially effective treatments and thus broaden its role from predictive assessment to actionable insights in patient management.


\section*{Methods}
\small

\subsection*{Data description}

\blktitle{Databases.} We used two large electronic health record (EHR) datasets: the Medical Information Mart for Intensive Care (MIMIC) IV and the eICU Collaborative Research Database. MIMIC-IV is a publicly available collection containing de-identified health-related data associated with over 70,000 patients staying in critical care units, and eICU consists of health data of over 200,000 patients from multiple centers in the United States.

\blktitle{Medical outcomes identification.} We focused on six critical care outcomes: acute kidney injury (AKI), acute respiratory distress syndrome (ARDS), circulatory failure, death, delirium, and sepsis, which are mostly frequently occurring cases in critical care and significantly impact patient prognosis. In both databases, we identified these outcomes using the following criteria: AKI was identified using the KDIGO criteria\cite{KDIGO2012}, ARDS using the Berlin criteria\cite{theardsdefinitiontaskforce*AcuteRespiratoryDistress2012}, circulatory failure by looking at the lactic acid level and MAP\cite{hylandEarlyPredictionCirculatory2020}, death using the hospital mortality flag, delirium using the CAM-ICU-related score, and sepsis using the SOFA score calculated in the MIMIC-IV-Derived dataset\cite{johnsonMIMICIVFreelyAccessible2023}. 
For details of our implementations of these criteria, please refer to the \textit{Supplements D.1}.

\blktitle{Time-series construction.} Before learning the prediction model, we organized the data of each patient into temporally structured series.  
Patient hospitalization data were chronologically arranged to create time-series, which are partitioned into non-overlapping 2-hour periods. 
The data from 14 days, i.e., a time window including 168 time points, can be used to predict the outcome in the next 24 hours.
We slide the 14-day time window at an interval of 2 hours to form sufficient prediction points and use all the available time windows during the whole hospitalization period to learn the outcomes prediction model and obtain an average of 74.42 prediction points for each patient. 
To address the missing observations, rather than using imputation methods, we preserved the integrity of the original observations by introducing a binary indicator for each variable, signaling whether data was missing. 
It is shown in \textit{Supplements Fig. S1} that the Missing rates of variables do not significantly bias causal discovery results, supporting our missing value preprocessing method.
For variables measured multiple times within the same time window, we take the most recent measurement. 

We also include three static variables, i.e., age, gender, and ethnicity, which are assumed to be influential factors for the outcomes and share the same causal discovery process as the time-series data since the outcomes cannot causally influence these variables.
In the implementation, these static variables are encoded with a standalone MLP before aggregating with the encoded time-series (see Extended Data Fig. \ref{fig:train}) and we omit them in the following discussions for simplicity.

\blktitle{Labeling.} For each prediction point, the label was set to be 1 if the outcome is identified within the next 24 hours, and 0 otherwise. The label is ambiguous if we cannot identify whether the outcome occurs, which might happen when key data is missing for this patient (e.g., no creatinine data for AKI prediction). These data with ambiguous labels are disregarded in the training process.

\subsection*{Causality in artificial intelligence}
Causal deep learning is an emerging machine-learning approach, which incorporates the causal relationships concealed within complex distributions to alleviate the strong data dependency of existing artificial intelligence techniques, and ensures stability and interpretability across different environments as well.  For more detailed literature reviews, please refer to  \textit{Supplements A}.

\blktitle{Structural causal model (SCM).}
Structural causal model (SCM) is a graphical representation of causal relationships among variables\cite{gongCausalDiscoveryTemporal2023, rungeCausalInferenceTime2023, assaadSurveyEvaluationCausal2022}, defined as a tuple $\mathcal{M} = \left\{ V, \mathcal{F} \right\}$, in which $V$ is the set of variables (in our setting, $V$ includes both clinical variables and outcomes) and $\mathcal{F}$ is the set of functions generating each variable. Elements in $\mathcal{F}$ are called structural equations and take the form $x_i = f_i(\text{Pa}(x_i), \epsilon_i)$ with $x_i$ being the variable, $\text{Pa}(x_i)$ the set of $x_i$'s causal parents, and $\epsilon_i$ the noise term.
Combining all variables and structural equations, we can represent SCM as a directed acyclic graph $\mathcal{G}=\left\{ V, E \right\}$, where $E$ is the set of directed edges. The causal effect of variable $x_i$ on variable $x_j$ is denoted as $x_i \rightarrow x_j$.

\blktitle{SCM in time-series.} When dealing with time-series data, we assume that causal effect does not flow backward in time and can construct a structural causal model that treats each observation of a time-series as a variable, i.e., full-time causal graph\cite{gongCausalDiscoveryTemporal2023}. However, constructing a full-time causal graph is extremely hard since we do not have sufficient dense time-series across the long time range. 
To address this issue, we utilize the window causal graph, which assumes time-invariant causal structures, i.e., the causal parents $\text{Pa}(x_{t, i})$ is irrelevant to time $t$. Further, a causal summary graph is a summary of the window causal graphs, with each node representing a time-series and edges indicating time-independent causal relations.

The task for causal discovery in time-series (also for our causal deep learning approach) is to identify the window causal graph and causal summary graph.
In \textit{Supplements A}, we briefly discuss existing causal discovery approaches.

\blktitle{Interventions.} An SCM allows for the description of interventions by replacing one of the structural equations in the system. Specifically,  we first use the do-operator $do(x_i = x_i')$ to represent the effect of applying intervention on the variable $x_i$ (i.e., setting $x_i$ to $x_i'$) and then replace the structural equation $x_i = f_i(\text{Pa}(x_i), \epsilon_i)$ with $x_i = x_i'$. In the context of prediction, we would also like to elaborate the concept of \textit{valid interventions}, which are interventions that do not change the causal parents of the outcome variable\cite{arjovskyInvariantRiskMinimization2020a, rojas-carullaInvariantModelsCausal2018}, i.e., $x_i \notin \text{Pa}(y), do(x_i = x_i')$ is a valid intervention.

\blktitle{Granger causality.}
Granger causality (GC), first proposed by Granger et al.\cite{grangerInvestigatingCausalRelations1969} in linear settings, is recently extended to nonlinear scenarios including neural networks \cite{marinazzoKernelgrangerCausalityAnalysis2008, tankNeuralGrangerCausality2022, loweAmortizedCausalDiscovery2022, chengCUTSNeuralCausal2023, schwabCXPlainCausalExplanations2019}. The basic idea is to test if each input ``helps'' the prediction $\hat{y}$. Specifically, we define the Granger causality as follows\cite{tankNeuralGrangerCausality2022, chengCUTSNeuralCausal2023}:
\begin{definition}
\label{granger_def}
Variable $x_i$ Granger cause $y$ if and only if there exists $x'_{i} \neq x_{i}$, 
\begin{equation}
    f_\theta(x_{1},... , x'_{i}, ..., x_{N}) \neq f_\theta(x_{1}, ...,  x_{i}, ..., x_{N})
\end{equation}
i.e., the input variable $x_i$ influences the prediction of $y$.
\end{definition}
GC can be efficiently inferred with LASSO or Group LASSO \cite{lozanoGroupedGraphicalGranger2009, basuRegularizedEstimationSparse2015} because the weight of non-contributing inputs can be minimized with $\mathcal{L}_1$ regularizer, which form sparse input. However, when applying GC to neural networks, the difficulty is that simple regularization on the network weight may not help to create sparse input, and thus may not help to infer Granger causality.

\subsection*{Outcomes prediction with causal discovery}

Let $x_{t, i}^p$ denote the input variables, $y_{j, p}$ the outcome, and $f_\Theta$ the neural network, we aim to learn a prediction model that can infer $y_{j, p}$ from $x_{t, i}^p$, where $t\in \{1, ..., T\}, i\in \{1, ..., N\}, j\in \{1, ..., M\}, p\in \{1, ..., P\}$ and $T$ is the time-window size of the input, $N$, $M$ and $P$ are respectively the number of input variables, outcomes, and patients. 
After representing the input variables as a matrix $\mathbf{X}_p = \left\{ x_{t, i}^p \right\}_{t=1, i=1}^{T-1, N-1}$ and the outcome as a vector $\mathbf{y}_p = \left\{ y_{j,p} \right\}_{j=1}^{M}$, we can formulate the learning problem as
\begin{equation}
    \min_{\Theta} \sum_{p=1}^{P} \mathcal{L} \left( f_\Theta(\mathbf{X}_p), \mathbf{y}_p \right),
\end{equation}
where $\mathcal{L}$ is the loss function, e.g., cross-entropy loss for binary classification, and $\Theta$ is the parameters of the neural network. To simplify the notation, we denote the network for predicting a single outcome $y_j$ as $f_{\theta_j}$, and $f_{\Theta}(\mathbf{X}_p) = \left\{ f_{\theta_j}(\mathbf{X}_p) \right\}_{j=0}^{M-1}$, $\mathcal{L}\left( f_{\theta_j}(\mathbf{X}_p), \mathbf{y}_p \right) = \sum_{j=0}^{M-1} \mathcal{L}\left( f_{\theta_j}(\mathbf{X}_p), y_{j, p} \right)$.

\blktitle{Network structure.} For the prediction neural network, 
we implemented $f_{\theta_j}$ as an encoder-decoder structure, i.e.,
\begin{equation}
    f_{\theta_j}\left(\mathbf{X}_p^{\text{Pa}(y_{j}; \mathcal{G}^{\text{v2o}})}\right) = \text{Dec}_j\left(\text{Enc}\left(\mathbf{X}_p^{\text{Pa}(y_{j}; \mathcal{G}^{\text{v2o}})}\right)\right),
\end{equation}
where $\text{Enc}(\cdot)$ encodes the input variables into a latent representation, and $\text{Dec}_j$ decodes them to predict the $j$th outcome $y_{j}$. Here we use a shared encoder but distinguished decoders and use subscript $j$ to index the decoders across different outcomes. The encoder is implemented with an attention network for dynamic data and an MLP for static data, while the decoder is an MLP, with the detailed structure shown in Extended Data Fig. \ref{fig:method}.

\blktitle{Causal discovery.} To achieve interpretable and generalizable prediction, we incorporate causal discovery into the learning process. Specifically, we aim to discover the structural causal model (SCM) from the input variables to outcomes, i.e., the window causal graph $\mathcal{G}$, and use the discovered causal graph to guide the learning process. 
In the medical outcomes prediction, it is reasonable to assume that the outcome is caused by a subset of input variables but the outcome cannot be the cause of either the input variables or future outcomes. 
As a result, the causal graph $\mathcal{G}$ can be split into two parts: the variable-to-outcome (V2O) graph $\mathcal{G}^{\text{v2o}}$ and the variable-to-variable (V2V) $\mathcal{G}^{\text{v2v}}$ graph, while the outcome-to-outcome and outcome-to-variable graph is assumed to be empty. 

For the V2O graph, we denote the causal parents of outcome $y_{j}$ as $\text{Pa}(y_{j}; \mathcal{G^{\text{v2o}}})$ and the input data matrix as $\mathbf{X}_p^{\text{Pa}(y_{j}^p; \mathcal{G^{\text{v2o}}})} = \left\{ x_{t, i}^p \right\}_{t=1, i\in \text{Pa}(y_{j}^p; \mathcal{G^{\text{v2o}}})}^{T}$, then learn a neural network that only takes causal variables as input by optimizing the following objective
\begin{equation}
    \min_{\theta_j, \mathcal{G}} \sum_{p=0}^{P} \sum_{j=0}^{M} \mathcal{L}_j \left( f_{\theta_j}\left(\mathbf{X}_p^{\text{Pa}(y_{j}; \mathcal{G}^{\text{v2o}})}\right), \mathbf{y}_p \right) + \mathcal{R}(\mathcal{G}^{\text{v2o}}).
\end{equation}
Here $\mathcal{L}$ is the loss function (see \textit{Supplements D.2} for details) and
$\mathcal{R}(\mathcal{G}^{\text{v2o}})$ is a regularization term that penalizes the neural network $f_{\theta_j}$ taking non-causal variables as input, which is defined as
\begin{equation}
    \mathcal{R}(\mathcal{G}^{\text{v2o}}) = \sum_{i=1, j=1}^{N, M}  \mathbb{I}(i \in \text{Pa}(y_{j}; \mathcal{G}^{\text{v2o}}))
\end{equation}
with $\mathbb{I}(\cdot)$ being the indicator function. 
In the implementation, we conducted optimization with two alternating steps, please see \textit{Supplements D.3} for details. 

Similarly, the optimization for the V2V graph is defined as
\begin{equation}
\label{eqn:v2vloss}
    \min_{\theta_j, \mathcal{G}^{\text{v2v}}} \sum_{p=1}^{P}\sum_{j=1}^{M} \mathcal{L}_j \left( f_{\theta_j}\left(\mathbf{X}_p^{\text{Pa}(x_{t, i}); \mathcal{G}^{\text{v2v}}}\right), \mathbf{X}_{t+1} \right) + \mathcal{R}(\mathcal{G}^{\text{v2v}}).
\end{equation}

\blktitle{Causal probability matrix.} 
We implemented the $\mathcal{R}(\mathcal{G}^{\text{v2o}})$ and $\mathcal{R}(\mathcal{G}^{\text{v2v}})$ in Eq. \ref{eqn:v2vloss} with causal probability graph. 
Since the regularization term contains binary variables, the optimization problem is extremely hard to solve. To address this issue, we propose to relax the binary variables to continuous ones, i.e., the causal probability graph. Specifically, we define the causal probability matrix as $\mathbf{M}_t^{\text{v2o}} = \left\{ p_{t, i, j} \right\}_{i, j=0}^{N-1, M-1}$ with $p_{i, j, t}$ denoting the probability of $x_{t, i}$ being $y_{j}$'s causal parent, then the regularization term is defined as
\begin{equation}
    \mathcal{R}(\mathcal{G}^{\text{v2o}}) = \sum_{i=0}^{N-1}\sum_{j=0}^{M-1}\sum_{t=0}^{\tau-1}  p_{t, i, j}.
\end{equation} 
Consequently, the causal parents of outcome $y_{j}$ are determined by sampling from the causal probability matrix $\mathbf{M}_t^{\text{v2o}}, t=0, ..., \tau-1$, i.e., $\text{Pa}(y_{j}; \mathcal{G}^{\text{v2o}}) = \left\{ i; s_{t,i}=1\right\}_{i=0}^{N-1}$, where $s_{t,i}$ is sampled from the distribution $\text{Ber}(p_{t, i, j})$. 

Because the sampling process is non-differentiable, we use the Gumbel-Softmax trick \cite{jangCategoricalReparameterizationGumbelSoftmax2017} to relax the sampling process, i.e.,
\begin{equation}
    s_{t,i} = \frac{\exp\left(\left(\log(p_{t, i, j}) + g_{t, i}\right)/\tau\right)}{\sum_{i=0}^{N-1}\exp\left(\left(\log(p_{t, i, j}) + g_{t, i}\right)/\tau\right)},
\end{equation}
where $g_{t, i}$ is the Gumbel noise and $\tau$ is the temperature parameter. In practice, the learning process consists of two alternating steps, one for optimizing the prediction neural network and the other for optimizing the causal probability matrix. We only use the Gumbel-Softmax trick in the latter step, and please refer to \textit{Supplements D.3} for details.

\blktitle{Cumulative window graph.} 
In medical outcomes prediction, it is reasonable to assume that the causal effect of near time points is stronger than far ones, so we propose to incorporate a cumulative window graph into the learning process to penalize the causal probability with longer time lags. Specifically, we define 
\begin{equation}
p_{i,j,t}^{\text{v2o}} = \sigma\left(\prod_{w=1}^{t'} q_{i,j,w}^{\text{v2o}}\right), p_{i,j,t}^{\text{v2v}} = \sigma\left(\prod_{w=1}^{t'} q_{i,j,w}^{\text{v2v}}\right),
\end{equation}
where $\sigma(\cdot)$ is the sigmoid function that maps the input to the range of $[0, 1]$. In practice, the input time window (168 time points) is segmented into 14 chunks, and time points within a chunk are allocated the same $p_{i,j,t'}^{\text{v2o}}$, i.e. $t'=\lfloor \frac{t}{12}\rfloor$, allowing a much smaller causal probability matrix.
The parameter $q_{i,j,w}$ is pursued by optimizing the following objective
\begin{equation}
    \min_{\theta, \mathbf{Q}^{\text{v2o}}} \sum_{p=0}^{P} \mathcal{L} \left( f_\theta\left(\mathbf{X}_p^{\text{Pa}(y_{j}; \mathcal{G}^{\text{v2o}})}\right), \mathbf{y}_p \right) + \mathcal{R}(\mathcal{G}^{\text{v2o}};f_\theta),
\end{equation}
in which the parameter set $\mathbf{Q}^{\text{v2o}}=\{q_{i,j,w}^{\text{v2o}}\}_{t,i,j=1}^{\tau,N,M}$. Similarly, the corresponding objective for the V2V graph is defined as
\begin{equation}
    \min_{\theta, \mathbf{Q}^{\text{v2v}}} \sum_{p=0}^{P} \mathcal{L} \left( f_\theta\left(\mathbf{X}_p^{\text{Pa}(x_{t, i}; \mathcal{G}^{\text{v2v}})}\right), \mathbf{X}_{t+1} \right) + \mathcal{R}(\mathcal{G}^{\text{v2v}};f_\theta).
\end{equation}

\subsection*{Interpretability}
\blktitle{Controlled direct effect.}
Based on the discovered V2V and V2O graphs, we can provide causal pathways of the outcomes, demonstrating how each variable influences the subsequent one, and finally the outcome. Moreover, we can calculate the controlled direct effect (CDE) of each variable on the outcomes, i.e.,
\begin{equation}
    \text{CDE}(x;i) = f_\theta(x_{1},... , x_{i}, ..., x_{N}) - f_\theta(x_{1}, ...,  x'_{i}, ..., x_{N})
\end{equation}
with $x_i'$ being the perturbance value of $x_i$, to serve as an explainable causal reasoning tool by providing a quantitative contribution of causal variables \cite{harradonCausalLearningExplanation2018, schwabCXPlainCausalExplanations2019}. 
To better serve clinical practice, we developed a visualization tool to show the causal graphs and CDE values. The tool is built with Vue.js\footnote{https://vuejs.org/} and Chart.js\footnote{https://www.chartjs.org/}, as exemplified in Fig. \ref{fig:intp} and more examples are shown in \textit{Supplements C.3}. 

\blktitle{Efficient calculation.} To calculate each CDE value, we need first to perform a full inference of the neural network, then perturb the variable of interest, and finally perform another full inference. This process includes redundant calculations and is too time-consuming for inferring thousands of CDE values. For acceleration,  we propose to utilize the causally-decoupled inference techniques, which only update a subset of the hidden layers of the neural network after perturbing a variable. Specifically, the calculation process is as
\begin{equation}
    \begin{aligned}
        &\text{Inference:}\quad f_\Theta(\mathbf{X}_p) = \left\{ \text{Dec}_j\left(\text{Enc}\left(\mathbf{X}^{\text{Pa}(y_{j}; \mathcal{G}^{\text{v2o}})}_p\right)\right) \right\}_{j=0}^{M-1};\\
        &\text{Perturbations:}\quad f_\Theta(\mathbf{X}_p') = \left\{ \text{Dec}_j\left(\text{Enc}\left(\mathbf{X}^{\text{Pa}(y_{j}; \mathcal{G}^{\text{v2o}})'}_p\right)\right) \right\}_{j=0}^{M-1}.
    \end{aligned}
\end{equation}
Here  $\mathbf{X'}$ denotes perturbing variable $x_{t,i}$ to $x_{t,i}'$,
which means $(N+1)M$ inferences are required for calculating the CDE values of $N$ variables on $M$ outcomes. By contrast, when referring to the causal graph, we only need to calculate the sub-network corresponding to the causal parents of the variable of interest, i.e.,
\begin{equation}
    \text{Perturbations}: f_\Theta(\mathbf{X}_p') = \left\{ \text{Dec}_j\left(\text{Enc}\left(...\right)\right) \right\}_{j\in \text{Desc} (x_{t, i}; \mathcal{G}^{\text{v2o}})},
\end{equation}
where $\text{Desc} (x_{t, i}; \mathcal{G}^{\text{v2v}})$ is $x_{t, i}$'s descendants in the V2V graph. As a result, only $M+\|\mathcal{G}^{\text{v2o}}\|$ inferences are needed for calculating the CDE values, where $\|\mathcal{G}^{\text{v2o}}\|$ is the total number of the causal parents for each outcome.

\subsection*{Generalizability}
The out-of-distribution data can be seen as a result of interventions that might break the spurious associations learned in the training datasets and degenerate the performance \cite{christiansenCausalFrameworkDistribution2021}. 
Since causal relationships are stable across different environments, incorporating causal discovery to identify direct causation instead of spurious associations might produce AI models robust to variable interventions, i.e., with good generalizability. 
Theoretically, considering only one outcome $y_j$, let the function 
\begin{equation}
f^*\left( \mathbf{X} \right) = \mathbb{E}\left(y_j | \text{Pa}(y_{j}; \mathcal{G}^{\text{v2o}})\right)
\end{equation}
we show in the following theorem that $f^*(\cdot)$ is the optimal prediction function, which is equivalent to Theorem 4 in~\cite{rojas-carullaInvariantModelsCausal2018}.

\begin{theorem}
\label{theorem1}
Assuming that the causal graph $\mathcal{G}^{\text{v2o}}$ is the true causal graph, and the causal parents of the outcome $y_j$ is $\text{Pa}(y_{j}; \mathcal{G}^{\text{v2o}})$, 
\begin{equation}
    f^{*} \in \underset{f \in \mathcal{C}^0}{\arg \min } \sup _{\mathbb{P}^T \in \mathcal{P}} \mathbb{E}_{\left(\mathbf{X}^T, y^T\right) \sim \mathbb{P}^T}\left(y^T-f\left(\mathbf{X}^T\right)\right)^2,
\end{equation}
where $\mathcal{C}^0$ is the set of continuous functions $\mathbb{R}^{N\times \tau} \rightarrow \mathbb{R}$,
and $\mathcal{P}$ is the set of all possible distributions of $\left(\mathbf{X}^T, Y^T\right)$.
\end{theorem}

\subsection*{Baseline models}
To demonstrate its advantageous generalizability, we compared cDEEP with several existing generalizable AI approaches, including Invariant Risk Minimization (IRM), Group Distributional Robust Optimization (GroupDRO), and VREx: IRM achieves invariant prediction by learning an optimal classifier that is invariant across all domains; VREx introduces a variance regularizer that penalizes the variance of the training risks; GroupDRO minimizes the risk of the worst-performing domain by dynamically adjusting the weights of the domain risks. For more detailed literature reviews, please refer to  \textit{Supplements A}.

\subsection*{Performance evaluation}
We used the AUROC and the Area Under the Precision-Recall Curve (AUPRC, or average precision) to evaluate cDEEP’s predictive performance in the experiments. AUPRC is especially valuable in imbalanced datasets because it focuses on the performance related to the positive class, providing a more informative measure when dealing with rare events.
During the calculation of AUROC and AUPRC, we only included positive and negative samples and excluded the ambiguous samples, along with 95\% confidence intervals using the bootstrap method. 
As for model calibration, we used the Brier score, which is a proper scoring rule for measuring the accuracy of probabilistic predictions.

\section*{Data availability}

Data used for training, validation, and testing are from the open-sourced MIMIC-IV (\url{https://physionet.org/content/mimiciv/}) and eICU (\url{https://eicu-crd.mit.edu/}) datasets. 

\section*{Code Availability}

All codes for data preprocessing, training, and testing will be released to GitHub upon acceptance of this paper. We host our web-based interpretation tool on \url{https://cdeep.icu/} and the source code for the website will also be released.

\bibliography{ref}

\begin{thebibliography}{10}
\urlstyle{rm}
\expandafter\ifx\csname url\endcsname\relax
  \def\url#1{\texttt{#1}}\fi
\expandafter\ifx\csname urlprefix\endcsname\relax\def\urlprefix{URL }\fi
\expandafter\ifx\csname doiprefix\endcsname\relax\def\doiprefix{DOI: }\fi
\providecommand{\bibinfo}[2]{#2}
\providecommand{\eprint}[2][]{\url{#2}}

\bibitem{prosperiCausalInferenceCounterfactual2020}
\bibinfo{author}{Prosperi, M.} \emph{et~al.}
\newblock \bibinfo{journal}{\bibinfo{title}{Causal inference and counterfactual
  prediction in machine learning for actionable healthcare}}.
\newblock {\emph{\JournalTitle{Nature Machine Intelligence}}}
  \textbf{\bibinfo{volume}{2}}, \bibinfo{pages}{369--375},
  \doiprefix\url{10.1038/s42256-020-0197-y} (\bibinfo{year}{2020}).

\bibitem{vollmerMachineLearningArtificial2020}
\bibinfo{author}{Vollmer, S.} \emph{et~al.}
\newblock \bibinfo{journal}{\bibinfo{title}{Machine learning and artificial
  intelligence research for patient benefit: 20 critical questions on
  transparency, replicability, ethics, and effectiveness}}.
\newblock {\emph{\JournalTitle{BMJ}}} \textbf{\bibinfo{volume}{368}},
  \bibinfo{pages}{l6927}, \doiprefix\url{10.1136/bmj.l6927}
  (\bibinfo{year}{2020}).

\bibitem{shamoutMachineLearningClinical2021}
\bibinfo{author}{Shamout, F.}, \bibinfo{author}{Zhu, T.} \&
  \bibinfo{author}{Clifton, D.~A.}
\newblock \bibinfo{journal}{\bibinfo{title}{Machine {{Learning}} for {{Clinical
  Outcome Prediction}}}}.
\newblock {\emph{\JournalTitle{IEEE Reviews in Biomedical Engineering}}}
  \textbf{\bibinfo{volume}{14}}, \bibinfo{pages}{116--126},
  \doiprefix\url{10.1109/RBME.2020.3007816} (\bibinfo{year}{2021}).

\bibitem{yangUnboxBlackboxMedical2022}
\bibinfo{author}{Yang, G.}, \bibinfo{author}{Ye, Q.} \& \bibinfo{author}{Xia,
  J.}
\newblock \bibinfo{journal}{\bibinfo{title}{Unbox the black-box for the medical
  explainable {{AI}} via multi-modal and multi-centre data fusion: {{A}}
  mini-review, two showcases and beyond}}.
\newblock {\emph{\JournalTitle{Information Fusion}}}
  \textbf{\bibinfo{volume}{77}}, \bibinfo{pages}{29--52},
  \doiprefix\url{10.1016/j.inffus.2021.07.016} (\bibinfo{year}{2022}).

\bibitem{lauritsenExplainableArtificialIntelligence2020}
\bibinfo{author}{Lauritsen, S.~M.} \emph{et~al.}
\newblock \bibinfo{journal}{\bibinfo{title}{Explainable artificial intelligence
  model to predict acute critical illness from electronic health records}}.
\newblock {\emph{\JournalTitle{Nature Communications}}}
  \textbf{\bibinfo{volume}{11}}, \bibinfo{pages}{3852},
  \doiprefix\url{10.1038/s41467-020-17431-x} (\bibinfo{year}{2020}).

\bibitem{caicedo-torresISeeUVisuallyInterpretable2019a}
\bibinfo{author}{{Caicedo-Torres}, W.} \& \bibinfo{author}{Gutierrez, J.}
\newblock \bibinfo{journal}{\bibinfo{title}{{{ISeeU}}: {{Visually}}
  interpretable deep learning for mortality prediction inside the {{ICU}}}}.
\newblock {\emph{\JournalTitle{Journal of Biomedical Informatics}}}
  \textbf{\bibinfo{volume}{98}}, \bibinfo{pages}{103269},
  \doiprefix\url{10.1016/j.jbi.2019.103269} (\bibinfo{year}{2019}).

\bibitem{steyaertMultimodalDataFusion2023}
\bibinfo{author}{Steyaert, S.} \emph{et~al.}
\newblock \bibinfo{journal}{\bibinfo{title}{Multimodal data fusion for cancer
  biomarker discovery with deep learning}}.
\newblock {\emph{\JournalTitle{Nature Machine Intelligence}}}
  \bibinfo{pages}{1--12}, \doiprefix\url{10.1038/s42256-023-00633-5}
  (\bibinfo{year}{2023}).

\bibitem{ethayarajhAttentionFlowsAre2021}
\bibinfo{author}{Ethayarajh, K.} \& \bibinfo{author}{Jurafsky, D.}
\newblock \bibinfo{title}{Attention {{Flows}} are {{Shapley Value
  Explanations}}}.
\newblock In \bibinfo{editor}{Zong, C.}, \bibinfo{editor}{Xia, F.},
  \bibinfo{editor}{Li, W.} \& \bibinfo{editor}{Navigli, R.} (eds.)
  \emph{\bibinfo{booktitle}{Proceedings of the 59th {{Annual Meeting}} of the
  {{Association}} for {{Computational Linguistics}} and the 11th
  {{International Joint Conference}} on {{Natural Language Processing}}
  ({{Volume}} 2: {{Short Papers}})}}, \bibinfo{pages}{49--54},
  \doiprefix\url{10.18653/v1/2021.acl-short.8} (\bibinfo{publisher}{Association
  for Computational Linguistics}, \bibinfo{address}{Online},
  \bibinfo{year}{2021}).

\bibitem{yanInterpretableMortalityPrediction2020}
\bibinfo{author}{Yan, L.} \emph{et~al.}
\newblock \bibinfo{journal}{\bibinfo{title}{An interpretable mortality
  prediction model for {{COVID-19}} patients}}.
\newblock {\emph{\JournalTitle{Nature Machine Intelligence}}}
  \textbf{\bibinfo{volume}{2}}, \bibinfo{pages}{283--288},
  \doiprefix\url{10.1038/s42256-020-0180-7} (\bibinfo{year}{2020}).

\bibitem{srivastavaDropoutSimpleWay2014}
\bibinfo{author}{Srivastava, N.}, \bibinfo{author}{Hinton, G.},
  \bibinfo{author}{Krizhevsky, A.}, \bibinfo{author}{Sutskever, I.} \&
  \bibinfo{author}{Salakhutdinov, R.}
\newblock \bibinfo{journal}{\bibinfo{title}{Dropout: {{A Simple Way}} to
  {{Prevent Neural Networks}} from {{Overfitting}}}}.
\newblock {\emph{\JournalTitle{Journal of Machine Learning Research}}}
  \textbf{\bibinfo{volume}{15}}, \bibinfo{pages}{1929--1958}
  (\bibinfo{year}{2014}).

\bibitem{kroghSimpleWeightDecay1991}
\bibinfo{author}{Krogh, A.} \& \bibinfo{author}{Hertz, J.}
\newblock \bibinfo{title}{A {{Simple Weight Decay Can Improve
  Generalization}}}.
\newblock In \emph{\bibinfo{booktitle}{Advances in {{Neural Information
  Processing Systems}}}}, vol.~\bibinfo{volume}{4}
  (\bibinfo{publisher}{Morgan-Kaufmann}, \bibinfo{year}{1991}).

\bibitem{arjovskyInvariantRiskMinimization2020a}
\bibinfo{author}{Arjovsky, M.}, \bibinfo{author}{Bottou, L.},
  \bibinfo{author}{Gulrajani, I.} \& \bibinfo{author}{{Lopez-Paz}, D.}
\newblock \bibinfo{title}{Invariant {{Risk Minimization}}},
  \doiprefix\url{10.48550/arXiv.1907.02893} (\bibinfo{year}{2020}).
\newblock \eprint{1907.02893}.

\bibitem{sagawa*DistributionallyRobustNeural2019}
\bibinfo{author}{Sagawa*, S.}, \bibinfo{author}{Koh*, P.~W.},
  \bibinfo{author}{Hashimoto, T.~B.} \& \bibinfo{author}{Liang, P.}
\newblock \bibinfo{title}{Distributionally {{Robust Neural Networks}}}.
\newblock In \emph{\bibinfo{booktitle}{International {{Conference}} on
  {{Learning Representations}}}} (\bibinfo{year}{2019}).

\bibitem{kruegerOutDistributionGeneralizationRisk2021}
\bibinfo{author}{Krueger, D.} \emph{et~al.}
\newblock \bibinfo{title}{Out-of-{{Distribution Generalization}} via {{Risk
  Extrapolation}} ({{REx}})}.
\newblock In \emph{\bibinfo{booktitle}{Proceedings of the 38th {{International
  Conference}} on {{Machine Learning}}}}, \bibinfo{pages}{5815--5826}
  (\bibinfo{publisher}{PMLR}, \bibinfo{year}{2021}).

\bibitem{thorsen-meyerDynamicExplainableMachine2020}
\bibinfo{author}{{Thorsen-Meyer}, H.-C.} \emph{et~al.}
\newblock \bibinfo{journal}{\bibinfo{title}{Dynamic and explainable machine
  learning prediction of mortality in patients in the intensive care unit: A
  retrospective study of high-frequency data in electronic patient records}}.
\newblock {\emph{\JournalTitle{The Lancet Digital Health}}}
  \textbf{\bibinfo{volume}{2}}, \bibinfo{pages}{e179--e191},
  \doiprefix\url{10.1016/S2589-7500(20)30018-2} (\bibinfo{year}{2020}).

\bibitem{hylandEarlyPredictionCirculatory2020}
\bibinfo{author}{Hyland, S.~L.} \emph{et~al.}
\newblock \bibinfo{journal}{\bibinfo{title}{Early prediction of circulatory
  failure in the intensive care unit using machine learning}}.
\newblock {\emph{\JournalTitle{Nature Medicine}}}
  \textbf{\bibinfo{volume}{26}}, \bibinfo{pages}{364--373},
  \doiprefix\url{10.1038/s41591-020-0789-4} (\bibinfo{year}{2020}).

\bibitem{batesHarnessingAISepsis2022}
\bibinfo{author}{Bates, D.~W.} \& \bibinfo{author}{Syrowatka, A.}
\newblock \bibinfo{journal}{\bibinfo{title}{Harnessing {{AI}} in sepsis care}}.
\newblock {\emph{\JournalTitle{Nature Medicine}}}
  \textbf{\bibinfo{volume}{28}}, \bibinfo{pages}{1351--1352},
  \doiprefix\url{10.1038/s41591-022-01878-0} (\bibinfo{year}{2022}).

\bibitem{demirjianPredictiveAccuracyPerioperative2022}
\bibinfo{author}{Demirjian, S.} \emph{et~al.}
\newblock \bibinfo{journal}{\bibinfo{title}{Predictive accuracy of a
  perioperative laboratory test--based prediction model for moderate to severe
  acute kidney injury after cardiac surgery}}.
\newblock {\emph{\JournalTitle{JAMA}}} \textbf{\bibinfo{volume}{327}},
  \bibinfo{pages}{956--964}, \doiprefix\url{10.1001/jama.2022.1751}
  (\bibinfo{year}{2022}).

\bibitem{thanMachineLearningPredict2019}
\bibinfo{author}{Than, M.~P.} \emph{et~al.}
\newblock \bibinfo{journal}{\bibinfo{title}{Machine {{Learning}} to {{Predict}}
  the {{Likelihood}} of {{Acute Myocardial Infarction}}}}.
\newblock {\emph{\JournalTitle{Circulation}}} \textbf{\bibinfo{volume}{140}},
  \bibinfo{pages}{899--909}, \doiprefix\url{10.1161/CIRCULATIONAHA.119.041980}
  (\bibinfo{year}{2019}).

\bibitem{castroCausalityMattersMedical2020}
\bibinfo{author}{Castro, D.~C.}, \bibinfo{author}{Walker, I.} \&
  \bibinfo{author}{Glocker, B.}
\newblock \bibinfo{journal}{\bibinfo{title}{Causality matters in medical
  imaging}}.
\newblock {\emph{\JournalTitle{Nature Communications}}}
  \textbf{\bibinfo{volume}{11}}, \bibinfo{pages}{3673},
  \doiprefix\url{10.1038/s41467-020-17478-w} (\bibinfo{year}{2020}).

\bibitem{rungeCausalInferenceTime2023}
\bibinfo{author}{Runge, J.}, \bibinfo{author}{Gerhardus, A.},
  \bibinfo{author}{Varando, G.}, \bibinfo{author}{Eyring, V.} \&
  \bibinfo{author}{{Camps-Valls}, G.}
\newblock \bibinfo{journal}{\bibinfo{title}{Causal inference for time series}}.
\newblock {\emph{\JournalTitle{Nature Reviews Earth \& Environment}}}
  \textbf{\bibinfo{volume}{4}}, \bibinfo{pages}{487--505},
  \doiprefix\url{10.1038/s43017-023-00431-y} (\bibinfo{year}{2023}).

\bibitem{pearlCausalityModelsReasoning2009}
\bibinfo{author}{Pearl, J.}
\newblock \emph{\bibinfo{title}{Causality: {{Models}}, {{Reasoning}} and
  {{Inference}}}} (\bibinfo{publisher}{Cambridge University Press},
  \bibinfo{address}{USA}, \bibinfo{year}{2009}), \bibinfo{edition}{2nd} edn.

\bibitem{vowelsDyaDagsSurvey2021}
\bibinfo{author}{Vowels, M.~J.}, \bibinfo{author}{Camgoz, N.~C.} \&
  \bibinfo{author}{Bowden, R.}
\newblock \bibinfo{title}{D'ya like dags? {{A}} survey on structure learning
  and causal discovery}, \doiprefix\url{10.48550/arXiv.2103.02582}
  (\bibinfo{year}{2021}).
\newblock \eprint{2103.02582}.

\bibitem{shenStableLearningSample2020}
\bibinfo{author}{Shen, Z.}, \bibinfo{author}{Cui, P.}, \bibinfo{author}{Zhang,
  T.} \& \bibinfo{author}{Kunag, K.}
\newblock \bibinfo{journal}{\bibinfo{title}{Stable {{Learning}} via {{Sample
  Reweighting}}}}.
\newblock {\emph{\JournalTitle{Proceedings of the AAAI Conference on Artificial
  Intelligence}}} \textbf{\bibinfo{volume}{34}}, \bibinfo{pages}{5692--5699},
  \doiprefix\url{10.1609/aaai.v34i04.6024} (\bibinfo{year}{2020}).

\bibitem{moraffahCausalInferenceTime2021}
\bibinfo{author}{Moraffah, R.} \emph{et~al.}
\newblock \bibinfo{journal}{\bibinfo{title}{Causal inference for time series
  analysis: Problems, methods and evaluation}}.
\newblock {\emph{\JournalTitle{Knowledge and Information Systems}}}
  \textbf{\bibinfo{volume}{63}}, \bibinfo{pages}{3041--3085},
  \doiprefix\url{10.1007/s10115-021-01621-0} (\bibinfo{year}{2021}).

\bibitem{chengCUTSNeuralCausal2023}
\bibinfo{author}{Cheng, Y.} \emph{et~al.}
\newblock \bibinfo{title}{{{CUTS}}: {{Neural Causal Discovery}} from
  {{Irregular Time-Series Data}}}.
\newblock In \emph{\bibinfo{booktitle}{The {{Eleventh International
  Conference}} on {{Learning Representations}}}} (\bibinfo{year}{2023}).

\bibitem{chengCUTSHighDimensionalCausal2024}
\bibinfo{author}{Cheng, Y.} \emph{et~al.}
\newblock \bibinfo{journal}{\bibinfo{title}{{{CUTS}}+: {{High-Dimensional
  Causal Discovery}} from {{Irregular Time-Series}}}}.
\newblock {\emph{\JournalTitle{Proceedings of the AAAI Conference on Artificial
  Intelligence}}} \textbf{\bibinfo{volume}{38}}, \bibinfo{pages}{11525--11533},
  \doiprefix\url{10.1609/aaai.v38i10.29034} (\bibinfo{year}{2024}).

\bibitem{lagemannDeepLearningCausal2023}
\bibinfo{author}{Lagemann, K.}, \bibinfo{author}{Lagemann, C.},
  \bibinfo{author}{Taschler, B.} \& \bibinfo{author}{Mukherjee, S.}
\newblock \bibinfo{journal}{\bibinfo{title}{Deep learning of causal structures
  in high dimensions under data limitations}}.
\newblock {\emph{\JournalTitle{Nature Machine Intelligence}}}
  \textbf{\bibinfo{volume}{5}}, \bibinfo{pages}{1306--1316},
  \doiprefix\url{10.1038/s42256-023-00744-z} (\bibinfo{year}{2023}).

\bibitem{johnsonMIMICIVFreelyAccessible2023}
\bibinfo{author}{Johnson, A. E.~W.} \emph{et~al.}
\newblock \bibinfo{journal}{\bibinfo{title}{{{MIMIC-IV}}, a freely accessible
  electronic health record dataset}}.
\newblock {\emph{\JournalTitle{Scientific Data}}}
  \textbf{\bibinfo{volume}{10}}, \bibinfo{pages}{1},
  \doiprefix\url{10.1038/s41597-022-01899-x} (\bibinfo{year}{2023}).

\bibitem{pollardEICUCollaborativeResearch2018}
\bibinfo{author}{Pollard, T.~J.} \emph{et~al.}
\newblock \bibinfo{journal}{\bibinfo{title}{The {{eICU Collaborative Research
  Database}}, a freely available multi-center database for critical care
  research}}.
\newblock {\emph{\JournalTitle{Scientific Data}}} \textbf{\bibinfo{volume}{5}},
  \bibinfo{pages}{180178}, \doiprefix\url{10.1038/sdata.2018.178}
  (\bibinfo{year}{2018}).

\bibitem{KDIGO2012}
\bibinfo{journal}{\bibinfo{title}{{{KDIGO}}}}.
\newblock {\emph{\JournalTitle{Kidney International Supplements}}}
  \textbf{\bibinfo{volume}{2}}, \bibinfo{pages}{1},
  \doiprefix\url{10.1038/kisup.2012.1} (\bibinfo{year}{2012}).

\bibitem{santacruzHighLowPositive2021}
\bibinfo{author}{Santa~Cruz, R.}, \bibinfo{author}{Villarejo, F.},
  \bibinfo{author}{Irrazabal, C.} \& \bibinfo{author}{Ciapponi, A.}
\newblock \bibinfo{journal}{\bibinfo{title}{High versus low positive
  end-expiratory pressure ({{PEEP}}) levels for mechanically ventilated adult
  patients with acute lung injury and acute respiratory distress syndrome}}.
\newblock {\emph{\JournalTitle{Cochrane Database of Systematic Reviews}}}
  (\bibinfo{year}{2021}).

\bibitem{basilePathophysiologyAcuteKidney2012}
\bibinfo{author}{Basile, D.~P.}, \bibinfo{author}{Anderson, M.~D.} \&
  \bibinfo{author}{Sutton, T.~A.}
\newblock \bibinfo{journal}{\bibinfo{title}{Pathophysiology of {{Acute Kidney
  Injury}}}}.
\newblock {\emph{\JournalTitle{Comprehensive Physiology}}}
  \textbf{\bibinfo{volume}{2}}, \bibinfo{pages}{1303--1353},
  \doiprefix\url{10.1002/cphy.c110041} (\bibinfo{year}{2012}).

\bibitem{matuschakOrganInteractionsAdult1988}
\bibinfo{author}{Matuschak, G.~M.} \& \bibinfo{author}{Rinaldo, J.~E.}
\newblock \bibinfo{journal}{\bibinfo{title}{Organ interactions in the adult
  respiratory distress syndrome during sepsis: Role of the liver in host
  defense}}.
\newblock {\emph{\JournalTitle{Chest}}} \textbf{\bibinfo{volume}{94}},
  \bibinfo{pages}{400--406} (\bibinfo{year}{1988}).

\bibitem{herreroLiverLungInteractions2020}
\bibinfo{author}{Herrero, R.} \emph{et~al.}
\newblock \bibinfo{journal}{\bibinfo{title}{Liver--lung interactions in acute
  respiratory distress syndrome}}.
\newblock {\emph{\JournalTitle{Intensive Care Medicine Experimental}}}
  \textbf{\bibinfo{volume}{8}}, \bibinfo{pages}{48},
  \doiprefix\url{10.1186/s40635-020-00337-9} (\bibinfo{year}{2020}).

\bibitem{trillo-alvarezAcuteLungInjury2011}
\bibinfo{author}{{Trillo-Alvarez}, C.} \emph{et~al.}
\newblock \bibinfo{journal}{\bibinfo{title}{Acute lung injury prediction score:
  {{Derivation}} and validation in a population-based sample}}.
\newblock {\emph{\JournalTitle{European Respiratory Journal}}}
  \textbf{\bibinfo{volume}{37}}, \bibinfo{pages}{604--609},
  \doiprefix\url{10.1183/09031936.00036810} (\bibinfo{year}{2011}).

\bibitem{writinggroupforthealveolarrecruitmentforacuterespiratorydistresssyndrometrialartinvestigatorsEffectLungRecruitment2017}
\bibinfo{author}{{Writing Group for the Alveolar Recruitment for Acute
  Respiratory Distress Syndrome Trial (ART) Investigators}}.
\newblock \bibinfo{journal}{\bibinfo{title}{Effect of {{Lung Recruitment}} and
  {{Titrated Positive End-Expiratory Pressure}} ({{PEEP}}) vs {{Low PEEP}} on
  {{Mortality}} in {{Patients With Acute Respiratory Distress Syndrome}}: {{A
  Randomized Clinical Trial}}}}.
\newblock {\emph{\JournalTitle{JAMA}}} \textbf{\bibinfo{volume}{318}},
  \bibinfo{pages}{1335--1345}, \doiprefix\url{10.1001/jama.2017.14171}
  (\bibinfo{year}{2017}).

\bibitem{rivielloHospitalIncidenceOutcomes2016}
\bibinfo{author}{Riviello, E.~D.} \emph{et~al.}
\newblock \bibinfo{journal}{\bibinfo{title}{Hospital {{Incidence}} and
  {{Outcomes}} of the {{Acute Respiratory Distress Syndrome Using}} the
  {{Kigali Modification}} of the {{Berlin Definition}}}}.
\newblock {\emph{\JournalTitle{American Journal of Respiratory and Critical
  Care Medicine}}} \textbf{\bibinfo{volume}{193}}, \bibinfo{pages}{52--59},
  \doiprefix\url{10.1164/rccm.201503-0584OC} (\bibinfo{year}{2016}).

\bibitem{wickPulseOximetryDiagnosis2022}
\bibinfo{author}{Wick, K.~D.}, \bibinfo{author}{Matthay, M.~A.} \&
  \bibinfo{author}{Ware, L.~B.}
\newblock \bibinfo{journal}{\bibinfo{title}{Pulse oximetry for the diagnosis
  and management of acute respiratory distress syndrome}}.
\newblock {\emph{\JournalTitle{The Lancet Respiratory Medicine}}}
  \textbf{\bibinfo{volume}{10}}, \bibinfo{pages}{1086--1098},
  \doiprefix\url{10.1016/S2213-2600(22)00058-3} (\bibinfo{year}{2022}).

\bibitem{morrisMetabolicAcidosisCritically2008}
\bibinfo{author}{Morris, C.~G.} \& \bibinfo{author}{Low, J.}
\newblock \bibinfo{journal}{\bibinfo{title}{Metabolic acidosis in the
  critically ill: {{Part}} 2. {{Causes}} and treatment}}.
\newblock {\emph{\JournalTitle{Anaesthesia}}} \textbf{\bibinfo{volume}{63}},
  \bibinfo{pages}{396--411}, \doiprefix\url{10.1111/j.1365-2044.2007.05371.x}
  (\bibinfo{year}{2008}).

\bibitem{marikHemodynamicParametersGuide2011}
\bibinfo{author}{Marik, P.~E.}, \bibinfo{author}{Monnet, X.} \&
  \bibinfo{author}{Teboul, J.-L.}
\newblock \bibinfo{journal}{\bibinfo{title}{Hemodynamic parameters to guide
  fluid therapy}}.
\newblock {\emph{\JournalTitle{Annals of Intensive Care}}}
  \textbf{\bibinfo{volume}{1}}, \bibinfo{pages}{1},
  \doiprefix\url{10.1186/2110-5820-1-1} (\bibinfo{year}{2011}).

\bibitem{liImmuneDysfunctionLeads2020}
\bibinfo{author}{Li, D.} \emph{et~al.}
\newblock \bibinfo{journal}{\bibinfo{title}{Immune dysfunction leads to
  mortality and organ injury in patients with {{COVID-19}} in {{China}}:
  Insights from {{ERS-COVID-19}} study}}.
\newblock {\emph{\JournalTitle{Signal Transduction and Targeted Therapy}}}
  \textbf{\bibinfo{volume}{5}}, \bibinfo{pages}{1--3},
  \doiprefix\url{10.1038/s41392-020-0163-5} (\bibinfo{year}{2020}).

\bibitem{adriePersistentLymphopeniaRisk2017}
\bibinfo{author}{Adrie, C.} \emph{et~al.}
\newblock \bibinfo{journal}{\bibinfo{title}{Persistent lymphopenia is a risk
  factor for {{ICU-acquired}} infections and for death in {{ICU}} patients with
  sustained hypotension at admission}}.
\newblock {\emph{\JournalTitle{Annals of Intensive Care}}}
  \textbf{\bibinfo{volume}{7}}, \bibinfo{pages}{30},
  \doiprefix\url{10.1186/s13613-017-0242-0} (\bibinfo{year}{2017}).

\bibitem{gouveabogossianEffectIncreasedPositive2023}
\bibinfo{author}{Gouvea~Bogossian, E.} \emph{et~al.}
\newblock \bibinfo{journal}{\bibinfo{title}{The effect of increased positive
  end expiratory pressure on brain tissue oxygenation and intracranial pressure
  in acute brain injury patients}}.
\newblock {\emph{\JournalTitle{Scientific Reports}}}
  \textbf{\bibinfo{volume}{13}}, \bibinfo{pages}{16657},
  \doiprefix\url{10.1038/s41598-023-43703-9} (\bibinfo{year}{2023}).

\bibitem{velissarisHypomagnesemiaCriticallyIll2015}
\bibinfo{author}{Velissaris, D.}, \bibinfo{author}{Karamouzos, V.},
  \bibinfo{author}{Pierrakos, C.}, \bibinfo{author}{Aretha, D.} \&
  \bibinfo{author}{Karanikolas, M.}
\newblock \bibinfo{journal}{\bibinfo{title}{Hypomagnesemia in {{Critically Ill
  Sepsis Patients}}}}.
\newblock {\emph{\JournalTitle{Journal of Clinical Medicine Research}}}
  \textbf{\bibinfo{volume}{7}}, \bibinfo{pages}{911--918},
  \doiprefix\url{10.14740/jocmr2351w} (\bibinfo{year}{2015}).

\bibitem{aliyuInterleukin6CytokineOverview2022}
\bibinfo{author}{Aliyu, M.} \emph{et~al.}
\newblock \bibinfo{journal}{\bibinfo{title}{Interleukin-6 cytokine: {{An}}
  overview of the immune regulation, immune dysregulation, and therapeutic
  approach}}.
\newblock {\emph{\JournalTitle{International Immunopharmacology}}}
  \textbf{\bibinfo{volume}{111}}, \bibinfo{pages}{109130},
  \doiprefix\url{10.1016/j.intimp.2022.109130} (\bibinfo{year}{2022}).

\bibitem{cohenLegalEthicalConcerns2017}
\bibinfo{author}{Cohen, I.~G.}, \bibinfo{author}{Amarasingham, R.},
  \bibinfo{author}{Shah, A.}, \bibinfo{author}{Xie, B.} \& \bibinfo{author}{Lo,
  B.}
\newblock \bibinfo{journal}{\bibinfo{title}{The {{Legal And Ethical Concerns
  That Arise From Using Complex Predictive Analytics In Health Care}}}}.
\newblock {\emph{\JournalTitle{Health Affairs}}}
  \doiprefix\url{10.1377/hlthaff.2014.0048} (\bibinfo{year}{2017}).

\bibitem{theardsdefinitiontaskforce*AcuteRespiratoryDistress2012}
\bibinfo{author}{{The ARDS Definition Task Force*}}.
\newblock \bibinfo{journal}{\bibinfo{title}{Acute {{Respiratory Distress
  Syndrome}}: {{The Berlin Definition}}}}.
\newblock {\emph{\JournalTitle{JAMA}}} \textbf{\bibinfo{volume}{307}},
  \bibinfo{pages}{2526--2533}, \doiprefix\url{10.1001/jama.2012.5669}
  (\bibinfo{year}{2012}).

\bibitem{gongCausalDiscoveryTemporal2023}
\bibinfo{author}{Gong, C.}, \bibinfo{author}{Yao, D.}, \bibinfo{author}{Zhang,
  C.}, \bibinfo{author}{Li, W.} \& \bibinfo{author}{Bi, J.}
\newblock \bibinfo{title}{Causal {{Discovery}} from {{Temporal Data}}: {{An
  Overview}} and {{New Perspectives}}} (\bibinfo{year}{2023}).
\newblock \eprint{2303.10112}.

\bibitem{assaadSurveyEvaluationCausal2022}
\bibinfo{author}{Assaad, C.~K.}, \bibinfo{author}{Devijver, E.} \&
  \bibinfo{author}{Gaussier, E.}
\newblock \bibinfo{journal}{\bibinfo{title}{Survey and evaluation of causal
  discovery methods for time series}}.
\newblock {\emph{\JournalTitle{Journal of Artificial Intelligence Research}}}
  \textbf{\bibinfo{volume}{73}}, \bibinfo{pages}{767--819},
  \doiprefix\url{10.1613/jair.1.13428} (\bibinfo{year}{2022}).

\bibitem{rojas-carullaInvariantModelsCausal2018}
\bibinfo{author}{{Rojas-Carulla}, M.}, \bibinfo{author}{Scholkopf, B.},
  \bibinfo{author}{Turner, R.} \& \bibinfo{author}{Peters, J.}
\newblock \bibinfo{journal}{\bibinfo{title}{Invariant {{Models}} for {{Causal
  Transfer Learning}}}}.
\newblock {\emph{\JournalTitle{Journal of Machine Learning Research}}}
  (\bibinfo{year}{2018}).

\bibitem{grangerInvestigatingCausalRelations1969}
\bibinfo{author}{Granger, C. W.~J.}
\newblock \bibinfo{journal}{\bibinfo{title}{Investigating causal relations by
  econometric models and cross-spectral methods}}.
\newblock {\emph{\JournalTitle{Econometrica}}} \textbf{\bibinfo{volume}{37}},
  \bibinfo{pages}{424--438}, \doiprefix\url{10.2307/1912791}
  (\bibinfo{year}{1969}).
\newblock \eprint{1912791}.

\bibitem{marinazzoKernelgrangerCausalityAnalysis2008}
\bibinfo{author}{Marinazzo, D.}, \bibinfo{author}{Pellicoro, M.} \&
  \bibinfo{author}{Stramaglia, S.}
\newblock \bibinfo{journal}{\bibinfo{title}{Kernel-granger causality and the
  analysis of dynamical networks}}.
\newblock {\emph{\JournalTitle{Physical review E}}}
  \textbf{\bibinfo{volume}{77}}, \bibinfo{pages}{056215}
  (\bibinfo{year}{2008}).

\bibitem{tankNeuralGrangerCausality2022}
\bibinfo{author}{Tank, A.}, \bibinfo{author}{Covert, I.},
  \bibinfo{author}{Foti, N.}, \bibinfo{author}{Shojaie, A.} \&
  \bibinfo{author}{Fox, E.~B.}
\newblock \bibinfo{journal}{\bibinfo{title}{Neural granger causality}}.
\newblock {\emph{\JournalTitle{IEEE Transactions on Pattern Analysis and
  Machine Intelligence}}} \textbf{\bibinfo{volume}{44}},
  \bibinfo{pages}{4267--4279}, \doiprefix\url{10.1109/TPAMI.2021.3065601}
  (\bibinfo{year}{2022}).

\bibitem{loweAmortizedCausalDiscovery2022}
\bibinfo{author}{L{\"o}we, S.}, \bibinfo{author}{Madras, D.},
  \bibinfo{author}{Zemel, R.} \& \bibinfo{author}{Welling, M.}
\newblock \bibinfo{title}{Amortized causal discovery: Learning to infer causal
  graphs from time-series data}.
\newblock In \emph{\bibinfo{booktitle}{Proceedings of the {{First Conference}}
  on {{Causal Learning}} and {{Reasoning}}}}, \bibinfo{pages}{509--525}
  (\bibinfo{publisher}{PMLR}, \bibinfo{year}{2022}).

\bibitem{schwabCXPlainCausalExplanations2019}
\bibinfo{author}{Schwab, P.} \& \bibinfo{author}{Karlen, W.}
\newblock \bibinfo{title}{{{CXPlain}}: {{Causal Explanations}} for {{Model
  Interpretation}} under {{Uncertainty}}}.
\newblock In \emph{\bibinfo{booktitle}{Advances in {{Neural Information
  Processing Systems}}}}, vol.~\bibinfo{volume}{32} (\bibinfo{publisher}{Curran
  Associates, Inc.}, \bibinfo{year}{2019}).

\bibitem{lozanoGroupedGraphicalGranger2009}
\bibinfo{author}{Lozano, A.~C.}, \bibinfo{author}{Abe, N.},
  \bibinfo{author}{Liu, Y.} \& \bibinfo{author}{Rosset, S.}
\newblock \bibinfo{title}{Grouped graphical {{Granger}} modeling methods for
  temporal causal modeling}.
\newblock In \emph{\bibinfo{booktitle}{Proceedings of the 15th {{ACM SIGKDD}}
  International Conference on {{Knowledge}} Discovery and Data Mining}},
  {{KDD}} '09, \bibinfo{pages}{577--586},
  \doiprefix\url{10.1145/1557019.1557085} (\bibinfo{publisher}{Association for
  Computing Machinery}, \bibinfo{address}{New York, NY, USA},
  \bibinfo{year}{2009}).

\bibitem{basuRegularizedEstimationSparse2015}
\bibinfo{author}{Basu, S.} \& \bibinfo{author}{Michailidis, G.}
\newblock \bibinfo{journal}{\bibinfo{title}{Regularized estimation in sparse
  high-dimensional time series models}}.
\newblock {\emph{\JournalTitle{The Annals of Statistics}}}
  \textbf{\bibinfo{volume}{43}}, \bibinfo{pages}{1535--1567},
  \doiprefix\url{10.1214/15-AOS1315} (\bibinfo{year}{2015}).

\bibitem{jangCategoricalReparameterizationGumbelSoftmax2017}
\bibinfo{author}{Jang, E.}, \bibinfo{author}{Gu, S.} \& \bibinfo{author}{Poole,
  B.}
\newblock \bibinfo{title}{Categorical {{Reparameterization}} with
  {{Gumbel-Softmax}}}, \doiprefix\url{10.48550/arXiv.1611.01144}
  (\bibinfo{year}{2017}).
\newblock \eprint{1611.01144}.

\bibitem{harradonCausalLearningExplanation2018}
\bibinfo{author}{Harradon, M.}, \bibinfo{author}{Druce, J.} \&
  \bibinfo{author}{Ruttenberg, B.}
\newblock \bibinfo{title}{Causal {{Learning}} and {{Explanation}} of {{Deep
  Neural Networks}} via {{Autoencoded Activations}}},
  \doiprefix\url{10.48550/arXiv.1802.00541} (\bibinfo{year}{2018}).
\newblock \eprint{1802.00541}.

\bibitem{christiansenCausalFrameworkDistribution2021}
\bibinfo{author}{Christiansen, R.}, \bibinfo{author}{Pfister, N.},
  \bibinfo{author}{Jakobsen, M.~E.}, \bibinfo{author}{Gnecco, N.} \&
  \bibinfo{author}{Peters, J.}
\newblock \bibinfo{title}{A causal framework for distribution generalization},
  \doiprefix\url{10.48550/arXiv.2006.07433} (\bibinfo{year}{2021}).
\newblock \eprint{2006.07433}.

\end{thebibliography}



\section*{Acknowledgement}
This work was supported by the National Natural Science Foundation of China (Nos. 61931012 and 62088102 to J. S.), as well as the National Key Research and Development Program of China (No. 2021ZD0140406) and the project of the Medical Engineering Laboratory of Chinese PLA General Hospital (Nos. 2022SYSZZKY20 and 2022SYSZZKY11 to K. H.). The funding agencies had no role in considering the study design or in the collection, analysis, interpretation of data, writing of the report, or decision to submit the article for publication.

\section*{Author contributions}
J. S., K. H., and Q. D. conceived this project. J. S. supervised this research. 
Y. C. designed the study.
Y. C. and X. S. implemented the algorithm, conducted the experiments, and wrote the manuscript with input from all authors. 
X. S. and Z. W. conducted the experiments.
Q. Z. provided the clinical insights. 
All authors contributed to the interpretation of the results and the writing of the manuscript.

\section*{Competing interests}
The authors declare no competing financial interests.

\section*{Materials and Correspondence}
Correspondence and material requests should be addressed to Jinli Suo.

\setcounter{figure}{0}
\captionsetup[figure]{labelfont={bf},name={Extended Data Fig.},labelsep=period}

\begin{figure*}[p]
\centering
\includegraphics[width=\linewidth, trim={{5.75in 18in 5.75in 2in}}, clip]{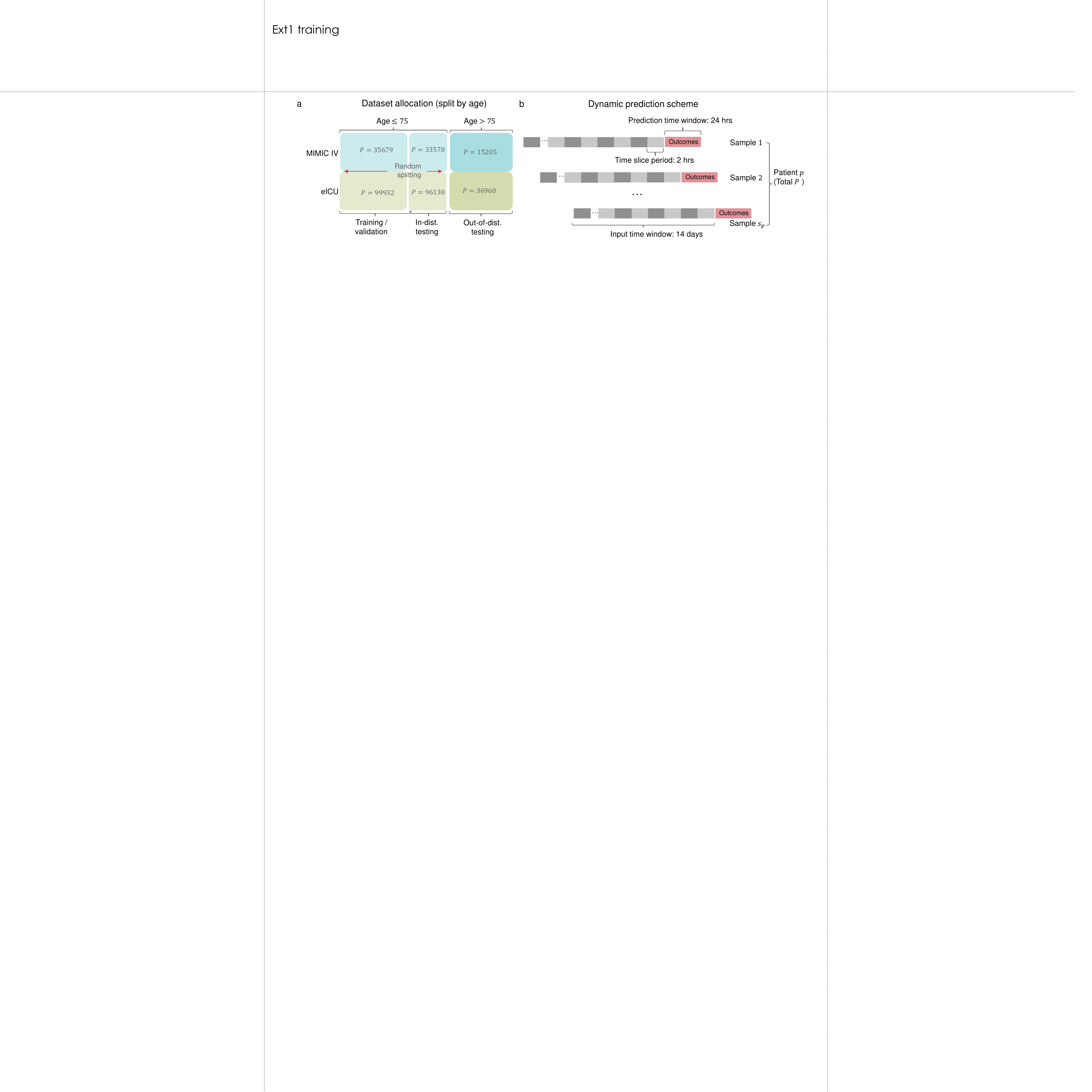}
\caption{ {\bf Detailed architecture of the algorithm.} \textbf{a,} Dataset allocation. We split the data from the MIMIC-IV and eICU databases into several subsets: training, validation, and in-distribution testing sets are randomly split from all patients with age $\leq 75$; out-of-distribution testing sets consist of all the patients with age $\geq 76$. $P$ denotes the number of patients in each subset.
\textbf{b,} Dynamic prediction scheme. Data from the electronic health record (EHR) are transformed into temporally structured sequences with each time slice being 2 hours. 
Taking the historical data from the preceding 14 days, the model predicts the patient's risk of developing a specific outcome at each time point in the next 24 hours, thereby generating multiple prediction ``samples''. Here $S_p$ denotes the number of samples for patient $p$.
}
\label{fig:dataalloc}
\end{figure*}

\begin{figure*}[p]
\centering
\includegraphics[width=\linewidth, trim={{5.75in 8.5in 5.75in 2in}}, clip]{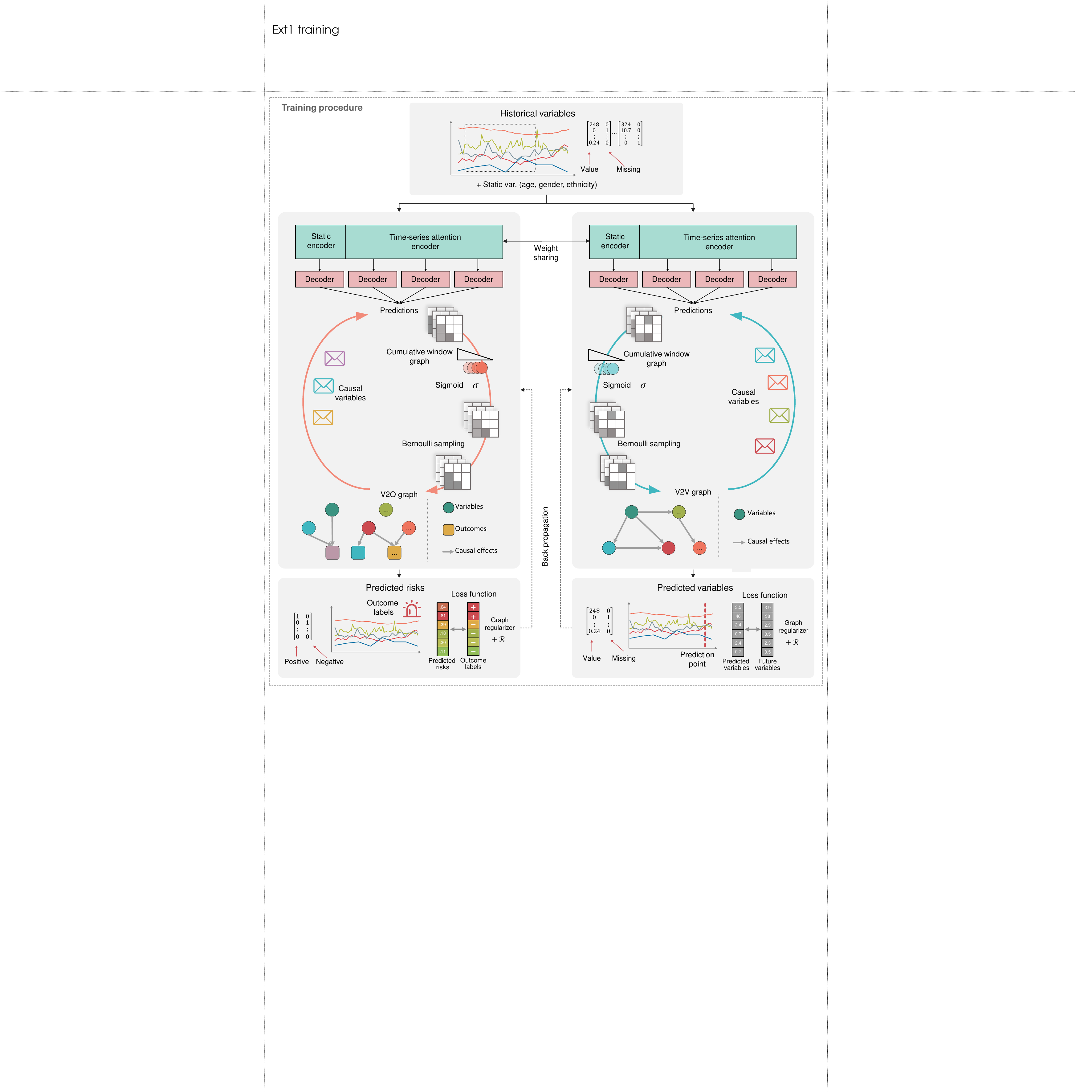}
\caption{{\bf Detailed architecture of the algorithm.} our algorithm, cDEEP, comprised two main modules---one predicts outcomes and the other decomposes the inference path of the prediction, optimizes the causal graphs (V2O and V2V graphs) and neural networks (outcomes and variable prediction models) iteratively to unveil direct causal relationships, enhancing the model's interpretability and generalizability. }
\label{fig:train}
\end{figure*}

\begin{figure*}[p]
    \centering
    \includegraphics[width=\linewidth, trim={{6.75in 13in 6.75in 2in}}, clip]{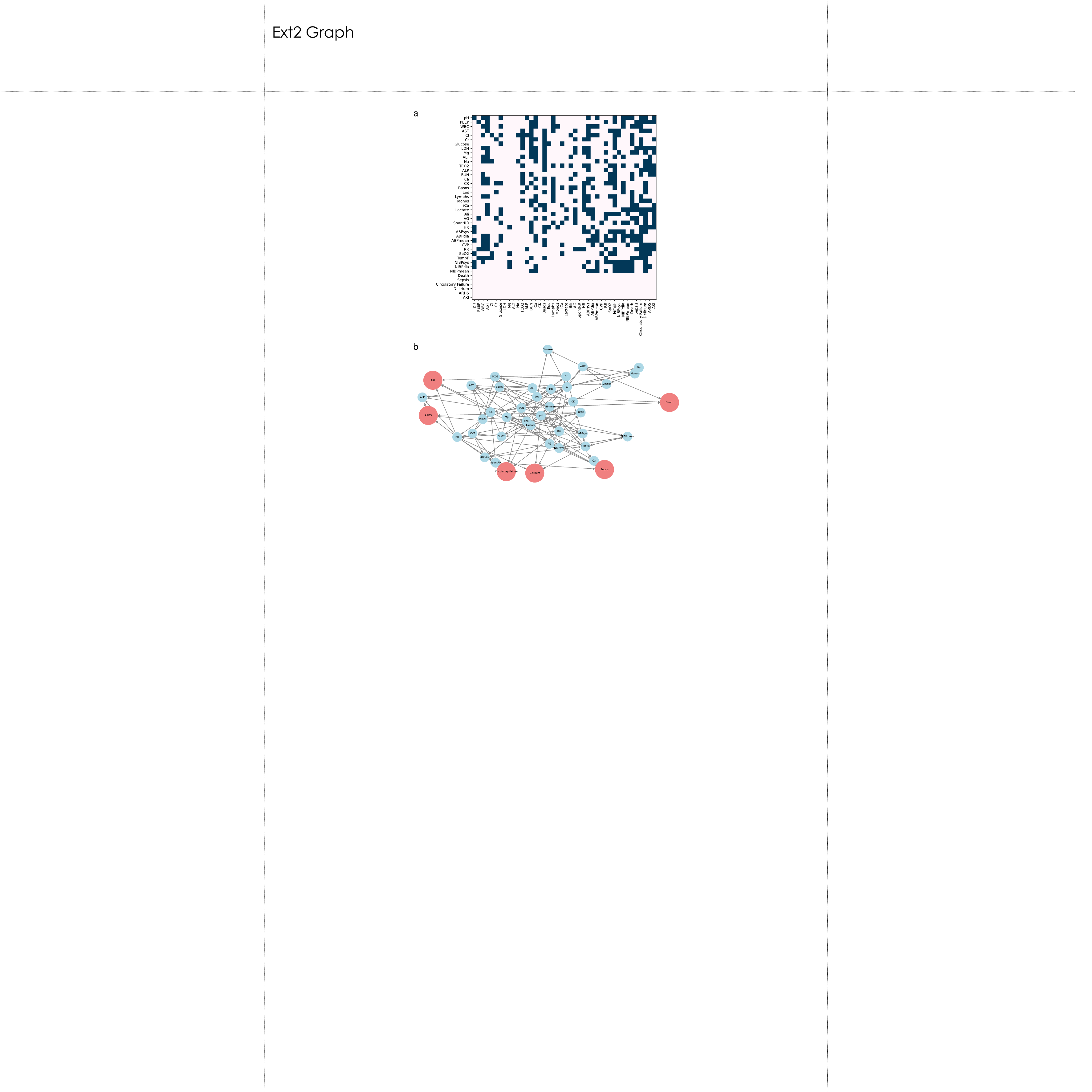}
    \caption{
        \textbf{Visualized causal graph of the input variables and outcomes.} 
        \textbf{a,} Thresholded causal probability matrix.
        \textbf{b,} Visualized causal graph. The nodes represent the variables, and the edges represent the causal relationships. For better visualization, we only show the most contributive causal relationships in the graph. Note that this graph is a summary causal graph, representing the causal relations without referring to time.
    }
    \label{fig:graph}
\end{figure*}

\end{document}